\documentclass{article}

\pdfoutput=1

\PassOptionsToPackage{numbers, sort&compress}{natbib}

     \usepackage[preprint]{neurips_2019}

\usepackage[utf8]{inputenc} 
\usepackage[T1]{fontenc}    
\usepackage{hyperref}       
\usepackage{url}            
\usepackage{booktabs}       
\usepackage{amsfonts}       
\usepackage{nicefrac}       
\usepackage{microtype}      

\usepackage{amsthm,amssymb, amsmath}
\usepackage{graphicx}

\usepackage{algorithmic}
\usepackage{algorithm}

\title{Robustness Against Outliers For Deep Neural Networks
 By Gradient Conjugate Priors}

\author{%
  Pavel~Gurevich\thanks{Peoples' Friendship University of Russia} \\
  Institute of Mathematics,\\
  Free University Berlin\\
  14195 Berlin, Germany\\
  \texttt{gurevichp@gmail.com} \\
   \And
   Hannes~Stuke\thanks{Equal contribution} \\
  Institute of Mathematics,\\
  Free University Berlin\\
  14195 Berlin, Germany\\
  \texttt{hannes.stuke@gmail.com} \\
}

\newtheorem{theorem}{Theorem}[section]

\newtheorem{lemma}{Lemma}[section]

\newtheorem{corollary}{Corollary}[section]
\newtheorem{condition}{Condition}[section]

\theoremstyle{definition}

\newtheorem{remark}{Remark}[section]

\newcommand{\ep}{\varepsilon}

\renewcommand{\phi}{{\varphi}}

\newcommand{\pc}{p_{\rm c}}
\newcommand{\pg}{p_{\rm g}}
\newcommand{\po}{p_{\rm o}}

\newcommand{\Vg}{V_{\rm g}}
\newcommand{\Vo}{V_{\rm o}}
\newcommand{\Vp}{V_{\rm p}}
\newcommand{\Vs}{V_{\rm St}}

\newcommand{\mc}{m_{\rm c}}
\newcommand{\mg}{m_{\rm g}}
\newcommand{\mo}{m_{\rm o}}

\renewcommand{\mp}{m_{\rm p}}

\newcommand{\muo}{\mu_{\rm o}}

\newcommand{\omegao}{\omega_{\rm o}}
\newcommand{\Cgo}{C_{\rm go}}
\newcommand{\Dgo}{D_{\rm go}}
\newcommand{\Gg}{G_{\rm g}}
\newcommand{\Go}{G_{\rm o}}

\newcommand{\by}{{\mathbf y}}

\newcommand{\btau}{{\boldsymbol{\tau}}}
\newcommand{\bmu}{{\boldsymbol{\mu}}}

\newcommand{\bbR}{{\mathbb R}}

\newcommand{\bbE}{{\mathbb E}}

\let\phi=\varphi

\begin{document}

\maketitle

\begin{abstract}
We analyze a new robust method for the reconstruction
of probability distributions of observed data in the presence of output outliers. It is based on a so-called gradient conjugate prior (GCP) network which outputs the parameters of a prior. By rigorously studying the dynamics of the GCP learning process, we derive an explicit formula for correcting the obtained variance of the marginal distribution and removing the bias caused by outliers in the training set. Assuming a Gaussian (input-dependent) ground truth distribution contaminated with a proportion $\varepsilon$ of outliers, we show that the fitted mean is in a $c e^{-1/\varepsilon}$-neighborhood of the ground truth mean and the corrected variance is in a $b\varepsilon$-neighborhood of the ground truth variance, whereas the uncorrected variance of the marginal distribution can even be infinite. We explicitly find $b$ as a function of the output of the GCP network, without a priori knowledge of the outliers (possibly input-dependent) distribution. Experiments with synthetic and real-world data sets indicate that the GCP network fitted with a standard optimizer outperforms other robust methods for regression.
\end{abstract}

\section{Introduction}

Development of methods robust against outliers in the observed data is an important direction of machine learning and statistics~\cite{HuberRonchetti2011}. One distinguishes between  {\em input outliers} (i.e., outliers $x$ in the input space)  and {\em output outliers} (i.e., wrongly labeled samples $y$). The former can potentially be detected both during fitting neural networks and when one predicts labels of new data samples. The latter are visible at the fitting stage only and significantly distort the approximate distribution one uses for predictions afterwards.
Bayesian neural networks and ensemble methods can naturally detect input outliers at the prediction stage by assigning high uncertainty to them~\cite{Pawlowski2017,Lakshminarayanan17}. In order to deal with input outliers at the fitting stage, one can use a covariate shift importance sampling~\cite{Sugiyama2007,Wen2014}, which assumes the knowledge of training and test distributions $p_{\rm train}(x)$ and $p_{\rm test}(x)$ of the input variable and downweights the samples with small ratios $p_{\rm test}(x)/p_{\rm train}(x)$.

We concentrate on how to mitigate the influence of {\em output} outliers at the fitting stage. We will estimate unknown mean and variance of labels\footnote{We denote random variables by bold letters and the arguments of their probability
distributions by the corresponding non-bold letters.} $\by\sim\pg(y|x)$ ({\em ground truth distribution}) in spite of contamination by an {\em outliers distribution} $\po(y|x)$. More specifically, we assume that the labels in the training set have Huber's {\em contaminated distribution}~\cite{Huber1964}
\begin{equation}\label{eqGroundTruth}
  \pc(y|x) = (1-\ep) \pg(y|x) + \ep \po(y|x),
\end{equation}
where $\ep\in[0,1)$ represents the proportion of outliers. Henceforth, we omit conditioning on $x$ for notational ease.  We assume throughout that the ground truth distribution $\pg(y)$ is {\em univariate Gaussian} with mean $\mg$ and variance $\Vg$, and we denote by $\mo$ and $\Vo$ the mean and variance of the outliers distribution $\po(y)$. We do {\em not} impose restrictions on $\po(y)$ except for a certain polynomial decay at infinity, see technical assumptions in Sec.~\ref{secBifurcation} and the supplement (Appendix~B).

The main contributions of this paper are as follows. {\bf 1.} We prove that outliers cause a qualitative change in the structure of the energy surfaces of the GCP network (analyzed in~\cite{GurHannesGCP} in the {\em absence} of outliers). Namely, outliers make a global minimum bifurcate from infinity to a finite value (Theorem~\ref{thSmallEpEquilibrium}). In turn, this renders the predictive distribution from Gaussian into Student's t, whose variance $\Vs$ may be significantly larger than the ground truth variance $\Vg$. {\bf 2.} We show how the knowledge of the above finite equilibrium allows one to reconstruct the ground truth mean $\mg$ and variance $\Vg$ (Theorems~\ref{thMeanExpClose} and~\ref{thVpDiff}).

Our experiments in Sec.~\ref{secExperiments} with synthetic and real-world data sets indicate that the GCP network, fitted with a standard optimizer (Adam in our case), outperforms other robust methods, particularly by properly estimating the ground truth variance.

\subsection{Main idea}\label{subsecMainIdea}
\begin{figure}[tb]
\begin{center}
\begin{minipage}{0.48\linewidth}
\includegraphics[width=0.7\linewidth]{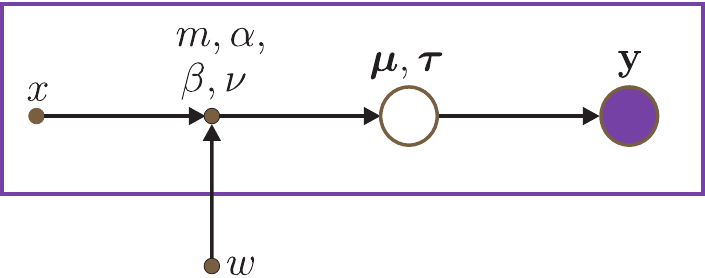}
\end{minipage}
\hfill
\begin{minipage}{0.48\linewidth}
\includegraphics[width=\linewidth]{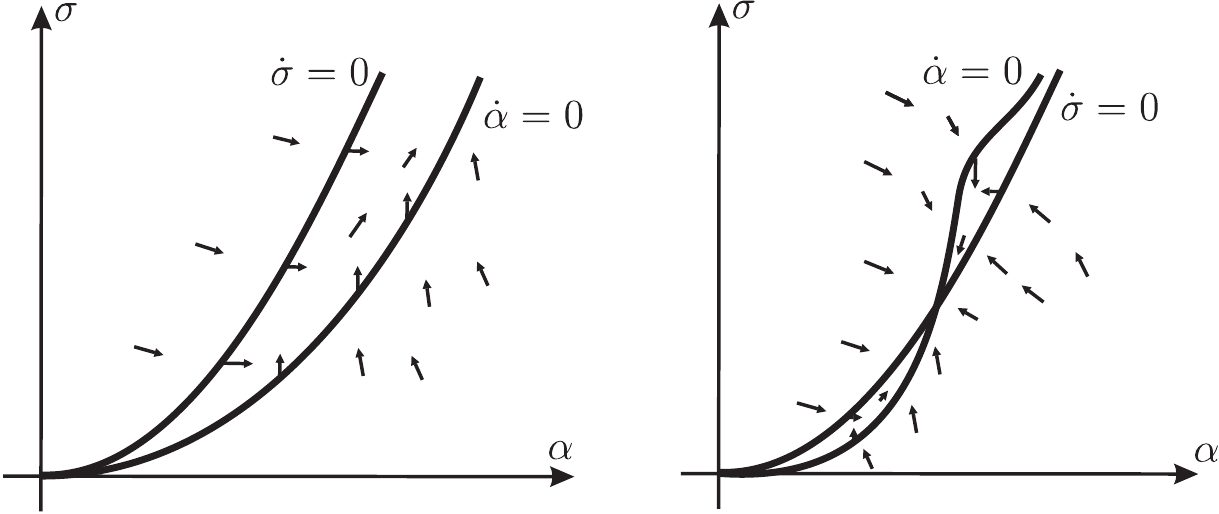}
\end{minipage}
\caption{Left: The graphical model of the GCP network with deterministic weights $w$. Right: Bifurcation of the vector field of~\eqref{eqDKalphaPure}, \eqref{eqDKsigmaPure} caused by outliers. The curves indicate the sets  $\dot\alpha=0$ and $\dot\sigma=0$. The arrows indicate directions of the vector field. Left: in the absence of outliers ($\ep=0$), $\alpha,\sigma\to\infty$. Right: in the presence of outliers (arbitrarily small $\ep>0$), an equilibrium bifurcates from infinity.}
\label{figGraphicalModelBifurcation}
\end{center}
\end{figure}

For each $x$ in the input space, we define a probabilistic model for a random variable $\by$ and latent variables $\bmu,\btau$
\begin{equation}\label{eqJointDistr}
p(y,\mu,\tau) = p(y|\mu,\tau)p(\mu,\tau|m,\nu,\alpha,\beta),
\end{equation}
where the likelihood $p(y|\mu,\tau)$ is assumed Gaussian with mean $\mu$ and precision $\tau$, while the latter are treated as random variables $\bmu,\btau$ with a normal-gamma distribution $p(\mu,\tau|m,\nu,\alpha,\beta)$. The parameters  $m,\nu,\alpha,\beta$ are functions of the input $x$, and are represented as outputs of multi-layer neural networks (Fig.~\ref{figGraphicalModelBifurcation}, left).
The marginal likelihood appears Student's t-distribution
\begin{gather}
t_{2\alpha}(y|m,\sigma/\alpha) = \int p(y,\mu,\tau)\,d\mu\, d\tau, \label{eqStudentst} \\
\sigma := \frac{\beta(\nu+1)}{\nu}.\label{eqSigma}
\end{gather}

In the standard Bayesian approach and $x$-independent case, one updates  $m,\nu,\alpha,\beta$ based on observations $y_1,\dots,y_N$. However, this is not possible in the neural networks framework because, on one hand, different $y_j$ belong to different input points $x_j$ and, on the other hand, one cannot update the outputs $m,\nu,\alpha,\beta$ of a neural network directly. The theory of Bayesian neural networks suggests to treat the weights of neural networks as random variables with a certain prior and to approximate their (usually analytically untractable) posterior~\cite{Neal95, Welling2011, Blundell2015, Kingma2015, GalGhahramani2015, HernandezLobato15,  LiuWang2016, LiGal2017}. Instead, we follow the gradient conjugate prior (GCP) method proposed in~\cite{GurHannesGCP}. We treat the weights $w$ of the neural networks as deterministic parameters. Given an observation $y_j$ corresponding to an input $x_j$, one can explicitly find the parameters $m',\nu',\alpha',\beta'$ of the posterior distribution of $\bmu,\btau$. We perform a gradient descent step towards minimization of the Kullback--Leibler (KL) divergence from the posterior to the prior, where the gradient is taken with respect to the weights $w$ of the neural networks representing $m,\nu,\alpha,\beta$. It is shown in~\cite{GurHannesGCP} that the GCP update is {\em equivalent} to maximizing the marginal log-likelihood $t_{2\alpha}(y|m,\sigma/\alpha)$. Furthermore, the above update of the weights $w$ induces an update of $m,\nu,\alpha,\beta$, which allows one to write a dynamical system (in the limit as the learning rate goes to $0$) for the evolution of $m,\nu,\alpha,\beta$ for each input $x$. This dynamical system takes the form
\begin{equation}\label{eqODE4}
\dot m = -\bbE\left[\frac{\partial K}{\partial m}\right],\quad \dot\alpha = -\bbE\left[\frac{\partial K}{\partial \alpha}\right],\quad
\dot\beta = -\bbE\left[\frac{\partial K}{\partial \beta}\right],\quad \dot\nu = -\bbE\left[\frac{\partial K}{\partial \nu}\right],
\end{equation}
where $K=K(m,\nu,\alpha,\beta)$ is the above KL-divergence, $\dot{}=d/dt$  stands for the derivative with respect to fictitious time $t$ and the expectations are taken with respect to the contaminated distribution $\pc(y)$ in~\eqref{eqGroundTruth}, see details in Sec.~\ref{secGCP}.

By analyzing system~\eqref{eqODE4}, we show that, for small $\ep>0$, the parameters $m(t),\alpha(t),\sigma(t)$ (see~\eqref{eqSigma}) converge to a finite equilibrium. We denote it by $m,\alpha,\sigma$ again (slightly abusing notation) and set
\begin{equation}\label{eqVp}
\mp:=m,\qquad \Vp := \dfrac{\sigma}{\alpha-A(\alpha)}.
\end{equation}
We call these quantities the {\em prognistic} mean and variance.\footnote{As opposed to the {\em predictive} variance $\Vs$ in~\eqref{eqStandardCPEstimate} of marginal Student's t-distribution $t_{2\alpha}(y|m,\sigma/\alpha)$ in~\eqref{eqStudentst}.} Here $A(\alpha)$ is monotone increasing from $0$ to $1$ and satisfies $\alpha-A(\alpha)>0$ for all $\alpha>0$, see  Fig.~4 in the supplement.
 It is defined as a unique root   $A=A(\alpha)$ of the equation
\begin{equation}\label{eqDKbetanu02}
 \frac{2\alpha+1}{\sqrt{2\pi}}\int\frac{y^2}{2(\alpha-A) + y^2}e^{-y^2/2} \,dy - 1 =0.
\end{equation}
Due to~Lemma~C.2 in the supplement, $A(\alpha)$ is well defined for all $\alpha>0$.   We show that, for small $\ep$, the prognostic mean $\mp$ is exponentially close  to $\mg$ (Theorem~\ref{thMeanExpClose}), while the prognostic variance $\Vp$ is linearly close to $\Vg$ (Theorem~\ref{thVpDiff}), namely,
\begin{equation}\label{eqFirstOrderCorrection}
\mg= \mp + O\bigl(e^{-c/\ep}\bigr),\quad \Vg=(1-b\ep)\Vp +  O(\ep^2),
\end{equation}
where $c>0$ and $b=b(\alpha)>0$ is defined in~(56) in the supplement. We emphasize the novelty of the prognostic variance $\Vp$ in~\eqref{eqVp}, which provides a correction of the usually used variance of the marginal distribution~\eqref{eqStudentst}. In our case, the latter is Student's t variance
\begin{equation}\label{eqStandardCPEstimate}
\Vs := \dfrac{\sigma}{\alpha-1}\ (\alpha>1),\quad  \Vs := \infty\ (\alpha\le 1).
\end{equation}
Note that $\Vs>\Vp$; moreover, $\Vs=\infty$ if $\alpha\le 1$. Therefore, even though Student's t-distribution is a popular choice in robust statistics and indeed provides a robust estimate of the mean, it significantly overestimates the ground truth variance in the presence of outliers, yielding an error $O(1)$. Our analysis  allows us to recover the ground truth variance via~\eqref{eqVp} up to an error $O(\ep)$ due to~\eqref{eqFirstOrderCorrection}.

A practical algorithm for fitting GCP networks is given in the supplement (Appendix~A).

\subsection{Related work}
There are several related approaches to mitigating the influence of output outliers.
One popular approach is based on fitting heavy-tailed distributions, such as Student's~t~\cite{Lange89,Lucas1996,Scheffler2008}. Effectively, our GCP method also fits Student's t-distribution, but additionally it reconstructs the ground truth variance $\Vg$ via~\eqref{eqVp} or~\eqref{eqFirstOrderCorrection}. Localization of a probabilistic model~\cite{WangBlei2015} generalizes heavy-tailed distributions. Localization principle allows the likelihood of each sample to depend on its own copy of a latent variable, while all the copies obey the same probability distribution. In particular cases, marginalizing the latent variables gives rise to Student's~t marginal likelihood. Another body of methods uses data reweighting. One can manually assign binary weights to samples~\cite{HuberRonchetti2011} or use the Bayesian framework~\cite{WangKucukelbirBlei2017}, in which the likelihood of each sample is raised to a power being a latent variable. The posterior of these latent variables is inferred together with the posterior of other latent variables in the model. Another type of reweighting is provided by so-called robust divergences, which are used instead of the Kullback--Leibler divergence either in directly approximating the ground truth distribution or in learning the posterior distribution of the parameters. For example, the $q$-entropy was used in~\cite{Ferrari2010}, while $\beta$- and $\gamma$-divergences were studied in~\cite{Basu1998,Gosh2016,Futami2017}. A number of papers develop robust gradient descent methods by detecting and reweighting the gradients of outliers during backpropagation~\cite{HollandIkeda2018,Prasad2018,Yin2018} or by removing outliers from a fitted model followed by refitting~\cite{Diakonikolas2018}. We emphasize that our GCP approach, in contrast to the above methods, can be trained in one run with any standard optimizer (such as Adam, RMSprop, etc.), and it does not require fine tuning additional hyperparameters or explicitly estimating the contamination proportion $\ep$. On the other hand, knowing $\ep$, one can reduce the error for the variance estimation to $O(\ep^2)$, see~\eqref{eqFirstOrderCorrection}.

\section{The GCP approach}\label{secGCP}

We recall the GCP approach introduced in~\cite{GurHannesGCP} and outlined in Sec.~\ref{subsecMainIdea}.

\subsection{GCP update}
We describe an update of $m,\nu,\alpha,\beta$ in~\eqref{eqJointDistr} directly, assuming $x$ to be fixed. We refer to Remark~\ref{remWGradient} and to~\cite{GurHannesGCP}   for details concerning an update of the weights $w$ of neural networks representing $m,\nu,\alpha,\beta$. Suppose we observe a new sample $ y $. Then, using the Bayes theorem, we find the conditional distribution of $(\bmu,\btau)$ under the condition that $\by= y $. This posterior distribution denoted by $p_{\rm post}(\mu,\tau)$ is also normal-gamma~\cite{BishopBook}, namely,
$
  p_{\rm post}(\mu,\tau) = p(\mu,\tau|m',\nu',\alpha',\beta'),
$
where the parameters are updated as follows:
\begin{equation}\label{eqCPUpdade}
m' = \dfrac{\nu m +  y }{\nu + 1},\quad
\nu' = \nu+1,\quad \alpha' = \alpha+\frac{1}{2},\quad
\beta' = \beta +   \frac{\nu}{\nu+1}\frac{( y - m)^2}{2}.
\end{equation}
However, in the framework of neural networks, one cannot update $m,\nu,\alpha,\beta$ directly. Instead, we {\em fix} $m',\nu',\alpha',\beta'$ according to~\eqref{eqCPUpdade} and use the  KL divergence from   $p_{\rm post}$ to $p$, see~\cite{Soch16}:
\begin{equation}\label{eqKLDivergence}
\begin{aligned}
& K( m,\nu,\alpha,\beta) :=   \frac{\alpha'(m-m')^2\nu}{2\beta'} +  \frac{\nu}{2\nu'} - \frac{1}{2}\ln\frac{\nu}{\nu'}-\frac{1}{2}\\
& - \alpha\ln\frac{\beta}{\beta'} + \ln\frac{\Gamma(\alpha)}{\Gamma(\alpha')} - (\alpha-\alpha')\Psi(\alpha') + \frac{\alpha'(\beta-\beta')}{\beta'},
\end{aligned}
\end{equation}
where $
 \Psi (\alpha):= {\Gamma '(\alpha)}/{\Gamma (\alpha)}
$ is the digamma function and $\Gamma(\alpha)$ is the gamma function.
After that, we update $m,\nu,\alpha,\beta$ by performing a gradient descent step in the direction $-\nabla K$. Recalling that the observations are sampled from the contaminated distribution $\pc(y)$, we can approximate the fitting process by the dynamical system~\eqref{eqODE4}.

\begin{remark}\label{remWGradient}
If $m,\nu,\alpha,\beta$ are parametrized by weights of neural networks, then the gradient of $K$ must be taken with respect to those weights, see the algorithm in the supplement (Appendix~A). The dynamics of the weights will induce a dynamics of $m,\nu,\alpha,\beta$ with the right-hand sides that contain the gradients of $m,\nu,\alpha,\beta$ with respect to the weights~\cite{GurHannesGCP}. However, they will enter as prefactors in~\eqref{eqODE4}. Hence any equilibrium of~\eqref{eqODE4} will be an equilibrium of the dynamical system for the weights.
\end{remark}

\subsection{Explicit dynamical system}
Dynamical system~\eqref{eqODE4} can be explicitly written as follows (cf.~(3.4)--(3.7) in~\cite{GurHannesGCP}):
\begin{align}
&\dot m  = (2\alpha+1)F(m,\sigma,\ep),\label{eqDKmPure}\\
&\dot \alpha  = -G(m,\sigma,\ep),\label{eqDKalphaPure}\\
& \dot \beta =  \frac{1}{\beta}H(m,\sigma,\ep),\quad
\dot\nu  =
-\frac{1}{\nu(\nu+1)}H(m,\sigma,\ep),\label{eqDKnuPure}
\end{align}
where
\begin{align}
& F(m,\sigma,\ep) := \int
\frac{z}{2\sigma  + z^2}\pc(y)\,dy,\label{eqDKmPureZero}\\
& G(m,\alpha,\sigma,\ep) :=  \int\ln\left(1+ \frac{z^2}{2\sigma}\right)\pc(y)\,dy + \Delta\Psi(\alpha),\label{eqDKalphaPureZero}\\
& H(m,\alpha,\sigma,\ep)  := \int\frac{\alpha z^2 - \sigma}{2\sigma + z^2}\pc(y)\,dy,\label{eqDKsigmaPureZero}
\end{align}
the integrals are taken over $\bbR$, $z=y-m$, $\pc(y)$ is defined in~\eqref{eqGroundTruth}, and
$\Delta\Psi(\alpha):=\Psi(\alpha)-\Psi(\alpha+1/2)$.
Equations~\eqref{eqDKnuPure} imply
\begin{equation}\label{eqDKsigmaPure}
 \dot\sigma = \left(\frac{(\nu+1)^2}{\nu^2\sigma}+\frac{\sigma}{(\nu+1)^2\nu^2}\right)H(m,\sigma,\alpha).
\end{equation}

The first goal of this paper is to show that, given outliers ($\ep>0$), fitting the parameters by the GCP method automatically yields finite values of $\alpha$ and $\sigma$. Theorem~\ref{thSmallEpEquilibrium} shows that finite $\alpha$ and $\sigma$ occur via bifurcation at infinity as $\ep$ becomes nonzero. The second goal is to show that the obtained prognostic mean $\mp$ and variance $\Vp$ in~\eqref{eqVp} do approximate the ground truth mean and variance in the sense of~\eqref{eqFirstOrderCorrection} given the output $m,\alpha,\beta,\nu$ of a fitted GCP network. This is done in Theorems~\ref{thMeanExpClose} and~\ref{thVpDiff}.

\section{Bifurcation of predictive distribution from Gaussian to Student's~t}\label{secBifurcation}

In this section, we show that an arbitrarily small percentage of outliers qualitatively changes the dynamics of~\eqref{eqDKmPure}--\eqref{eqDKnuPure}, \eqref{eqDKsigmaPure}, namely, it makes $\alpha$ and $\sigma$ converge to {\em finite} values. This changes the predictive distribution from Gaussian to Student's t. We will prove that this happens via bifurcation of the equilibrium $\alpha,\sigma$ at infinity (Fig.~\ref{figGraphicalModelBifurcation}, right). In the next section, we show that the correction given by the prognostic variance~$\Vp$ in~\ref{eqVp} is $\ep$-close to the ground truth variance $\Vg$.

Denote by $\muo^{(k)}$ the $k$th central moment of $\po(y)$. The following technical assumption requires that the mean $\mo$ or the variance $\Vo$ of outliers be large enough, or $\po(y)$ have heavy tails. It is used only in this section and does not depend on $\ep$.

\begin{condition}\label{condAnyPgSmallEps}
The outliers distribution $\po(y)$ satisfies $\Cgo>0$, where
\begin{equation}\label{eqCgo}
 \Cgo  :=(\mo-\mg)^4 + 6(\Vo-\Vg)(\mo-\mg)^2
 +3 (\Vo-\Vg)^2
 + (\muo^{(4)}-3\Vo^2) + 4\muo^{(3)}(\mo-\mg).
\end{equation}
\end{condition}

We will see that $\Cgo$ plays a role of an {\em indicator} of outliers. The larger $\Cgo$ is compared with $\Vg$, the better the GCP method recognizes samples from $\po(y)$ as outliers and the better it filters them out. A similar role of an indicator will be played by the absolute value of the constant
\begin{equation}\label{eqDgo}
\Dgo := (\mo-\mg)^3+3(\Vo-\Vg)(\mo-\mg) + \muo^{(3)}.
\end{equation}

%
%

\begin{theorem}\label{thSmallEpEquilibrium}
Let Condition~\ref{condAnyPgSmallEps} hold with some $\mg,\Vg,\mo,\Vo$. Then, for all sufficiently small $\ep>0$,
there exists a unique equilibrium $m_\ep,\alpha_\ep,\sigma_\ep$ of system~\eqref{eqDKmPure}, \eqref{eqDKalphaPure}, \eqref{eqDKsigmaPure}. The following asymptotics is true as $\ep\to 0$:
\begin{equation}\label{eqSmallEpEquilibrium2}
  m_\ep= \mg+ (\mo-\mg)\ep - \frac{\Cgo\Dgo}{6\Vg^3}\ep^2 + O(\ep^3),
\end{equation}
\begin{equation}\label{eqSmallEpEquilibrium1}
   \alpha_\ep=\frac{3\Vg^2}{\Cgo}\ep^{-1}+O(\ep^{-2}),\quad \sigma_\ep = \frac{3\Vg^3}{\Cgo}\ep^{-1} + O(\ep^{-2}).
\end{equation}
\end{theorem}

Theorem~\ref{thSmallEpEquilibrium} is proved in the supplement.

\section{Prognostic mean and variance}\label{subsecMeanExpClose}

The main practical question we answer in this section is the following. Given a finite equilibrium $(m,\alpha,\sigma)$ (as observed after the model is fitted), what can we tell about the ground truth mean $\mg$ and variance~$\Vg$? Due to~\eqref{eqVp}, the equilibrium $(m,\alpha,\sigma)$ uniquely determines prognostic mean $\mp$ and variance $\Vp$. Thus, for each $\ep$, there remain 4 unknowns $\mo,\Vo,\mg,\Vg$ in the 3 equations $F=G=H=0$ (see~\eqref{eqDKmPureZero}--\eqref{eqDKsigmaPureZero}). In this section, we assume they are functions of $\ep$ and obtain their asymptotics for small $\ep$ under the following condition.

\begin{condition}\label{condmoVoConst}
  Either $\mo(\ep)$ or $\Vo(\ep)$ is constant in $\ep$.
\end{condition}
  
The next theorem shows that  the prognostic mean $\mp$ is {\em exponentially} close to $\mg$.

\begin{theorem}\label{thMeanExpClose}
Let $\po(y):=\frac{1}{\sqrt{\Vo}}\tilde\po\left(\frac{y-\mo}{\sqrt{\Vo}}\right)$, where $\tilde\po(y)$ is an arbitrary distribution with zero mean and unit variance. Let $(m,\alpha,\sigma)$ be an equilibrium (independent of $\ep$) for system~\eqref{eqDKmPure}, \eqref{eqDKalphaPure}, \eqref{eqDKsigmaPure}. Let $\Vg(\ep)$ be bounded for all small $\ep$. Then there is an equilibrium $\mg(\ep)$ of Eq.~\eqref{eqDKmPure} such that
\begin{equation}\label{eqMeanExpClose1}
    \mg = \mp + O\big(e^{-c/\ep}\big)\quad\text{as } \ep\to 0
\end{equation}
for some $c>0$ that does not depend on $\ep$ and $\mg$.
\end{theorem}

The proof is given in the supplement (Appendix~C).\footnote{Theorem~\ref{thMeanExpClose} is proved under the assumption that either $|\mo(\ep)|$ or $\Vo(\ep)$ is bounded for small $\ep$, which is weaker than Condition~\ref{condmoVoConst}.}


Next, we analyze how much the prognostic variance $\Vp$ in~\eqref{eqVp}
differs from the ground truth variance $\Vg$.
 Theorem~\ref{thMeanExpClose} shows that  the equilibrium $m=\mp$ of~\eqref{eqDKmPure} is exponentially close to $\mg$. Therefore, to simplify our next statement and the technicalities of its proof, we assume that $m=\mg$.

\begin{theorem}\label{thVpDiff}
Let $\po(y):=\frac{1}{\sqrt{\Vo}}\tilde\po\left(\frac{y-\mo}{\sqrt{\Vo}}\right)$, where $\tilde\po(y)$ is an arbitrary distribution with zero mean and unit variance. Let $(\alpha,\sigma)$ be an equilibrium (independent of $\ep$) for system~\eqref{eqDKalphaPure}, \eqref{eqDKsigmaPure} with $m=\mg$. Then
\begin{equation}\label{eqVgDiffEp1}
  \Vg  = (1-b\ep)\Vp + O(\ep^2),
\end{equation}
where $b=b(\alpha)>0$ is defined in~$(56)$ in the supplement.
%
\end{theorem}

The proof is given in the supplement (Appendix~D). Moreover, we prove therein that any finite $(\alpha,\sigma)$ is realizable as an equilibrium for some $\mg,\Vg,\mo,\Vo,\ep$.

Asymptotics~\eqref{eqVgDiffEp1} should be compared with Student's t variance $\Vs$ in~\eqref{eqStandardCPEstimate}, which yields an error of order $1$ if $\alpha>1$ and an infinite error if $\alpha\le 1$.



\section{Experiments}\label{secExperiments}

\subsection{Methods}\label{subsecMethods}

\begin{figure*}[t]
\vskip 0.2in
\begin{center}
	\begin{minipage}{0.30\textwidth}
	   \includegraphics[width=\textwidth]{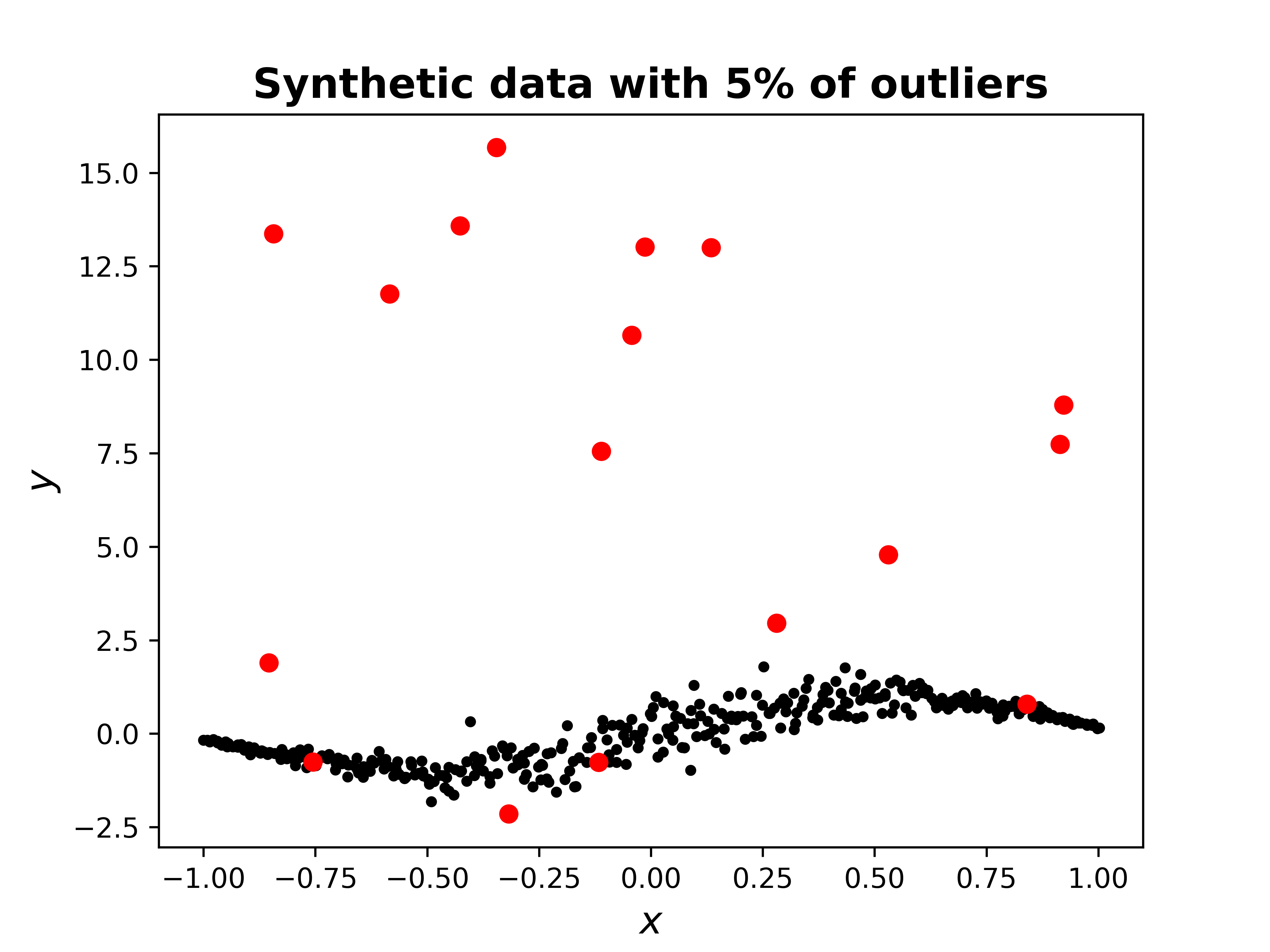}
	\end{minipage}
	\begin{minipage}{0.28\textwidth}
       \includegraphics[width=\textwidth]{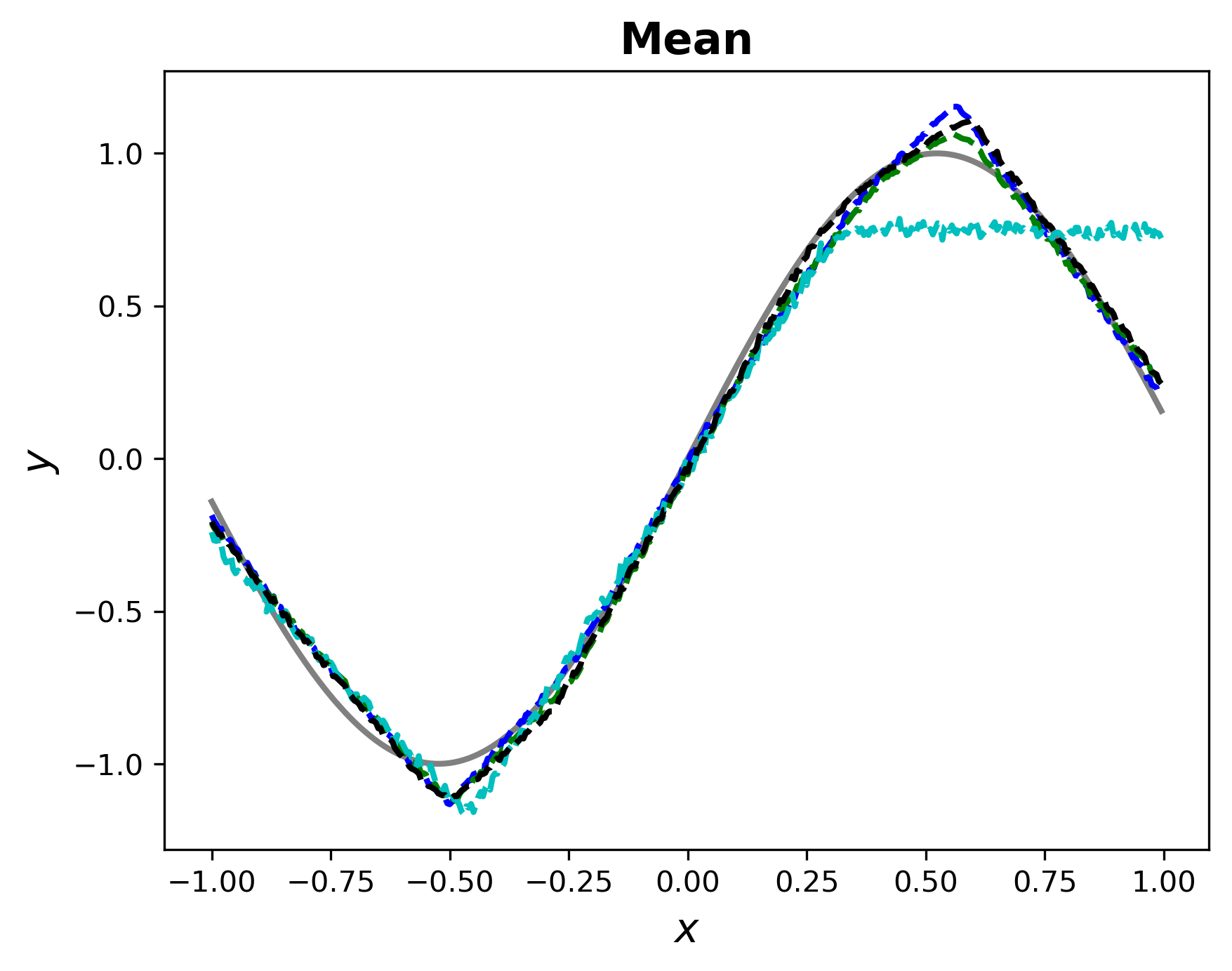}
	\end{minipage}
	\begin{minipage}{0.36\textwidth}
       \includegraphics[width=\textwidth]{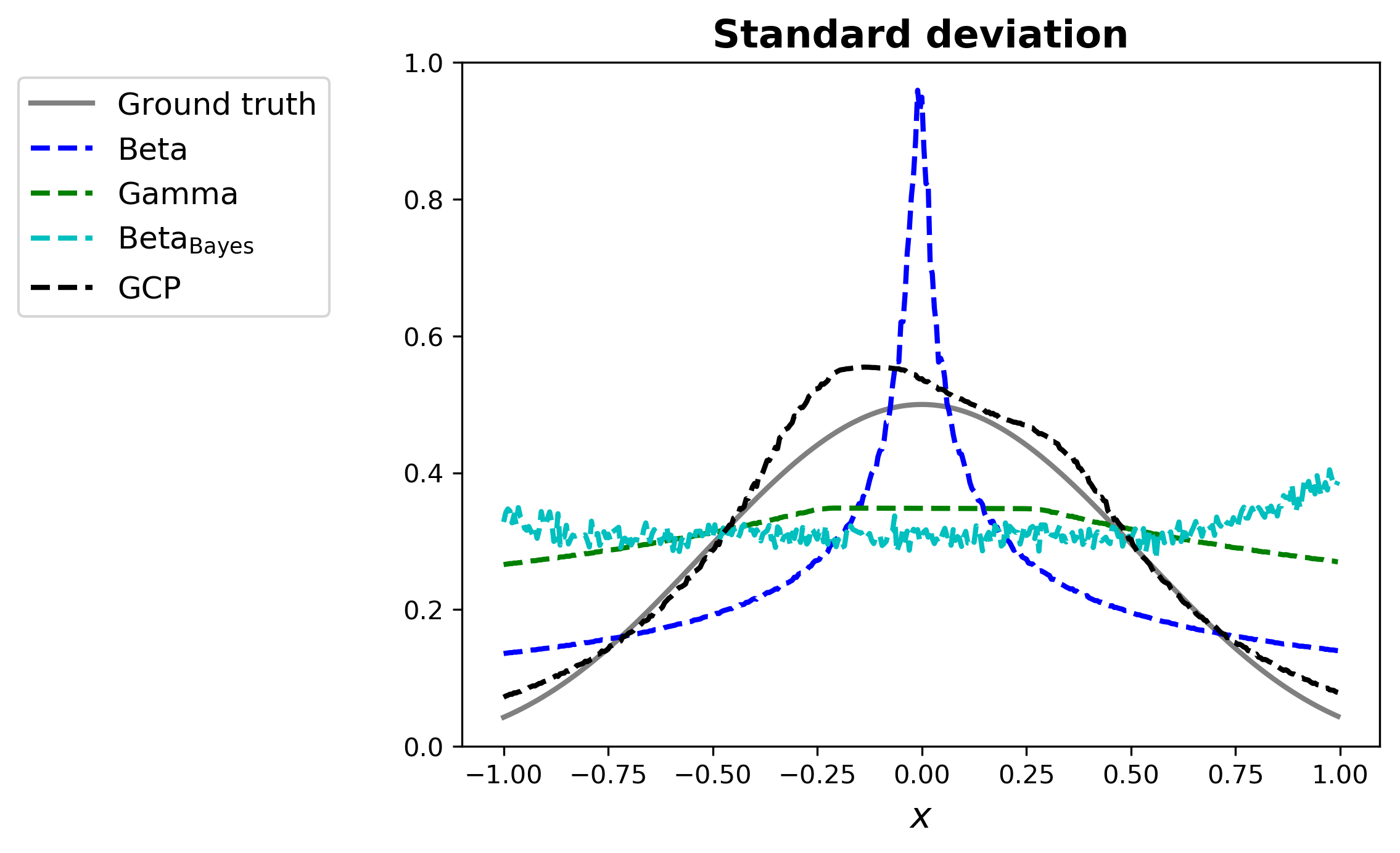}
	\end{minipage}
\caption{Left: Synthetic data complemented by 5\% of outliers (red disks). Middle: The means. Right: The standard deviations. The plots of GCP$_{\rm St}$ are omitted because the fitted variance $\Vs$ given by~\eqref{eqStandardCPEstimate} is either too large or infinite (due to small $\alpha$).}\label{figSyntheticDataWithOutliers}
\end{center}
\vskip -0.2in
\end{figure*}

\begin{figure*}[t]
\vskip 0.2in
\begin{center}
	\begin{minipage}{0.3\textwidth}
	   \includegraphics[width=\textwidth]{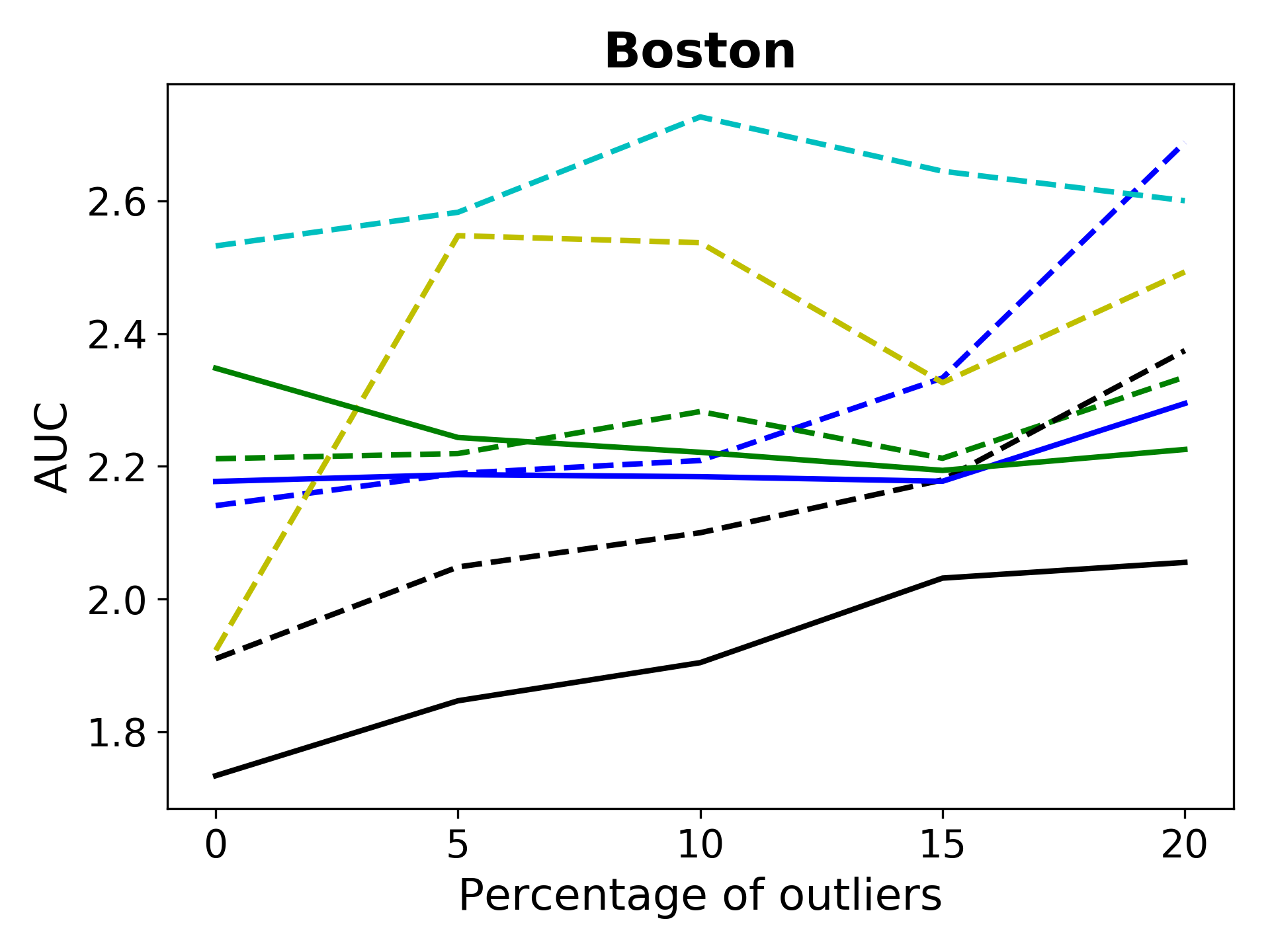}
	\end{minipage}
\hfill
	\begin{minipage}{0.3\textwidth}
       \includegraphics[width=\textwidth]{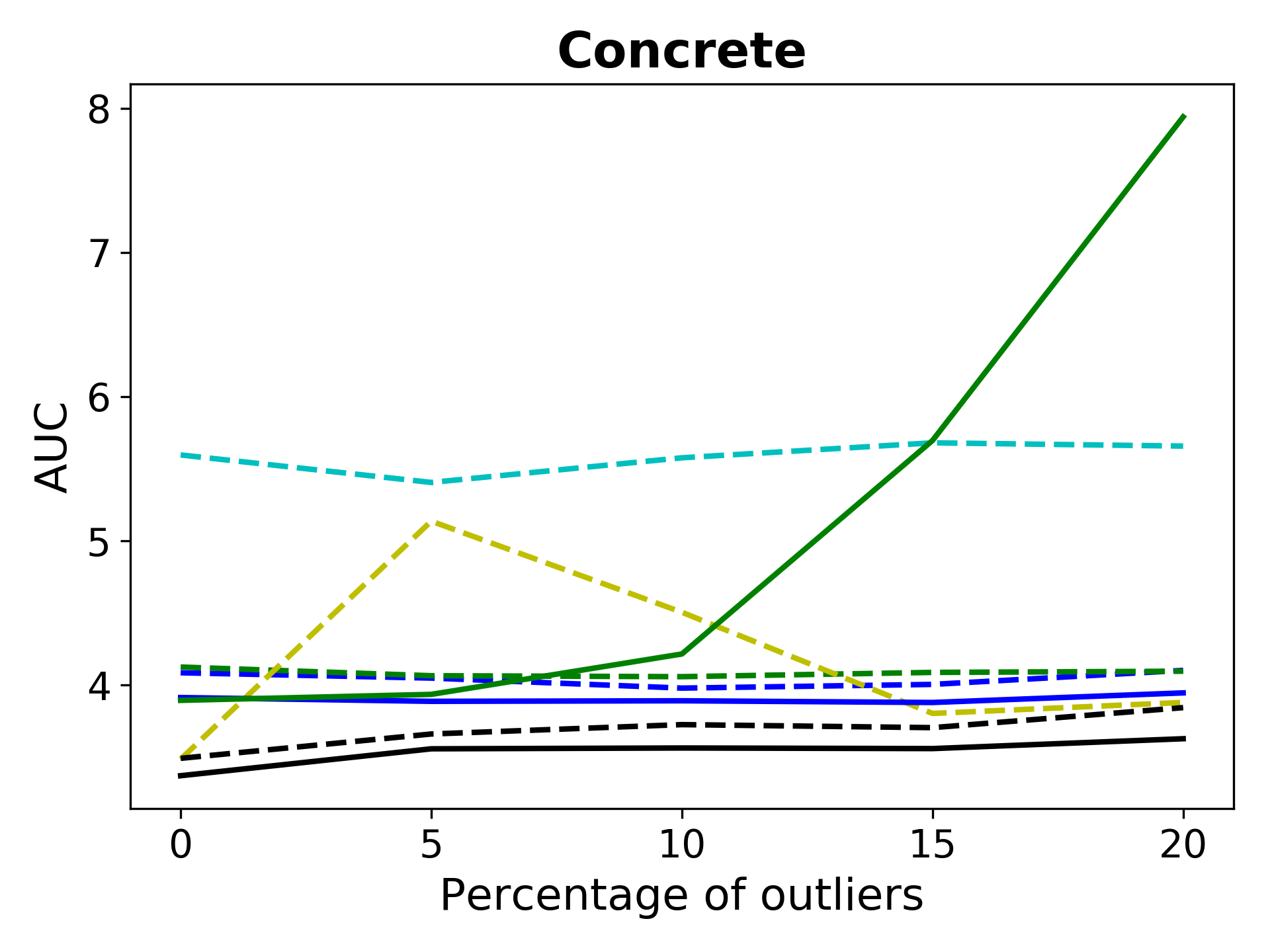}
	\end{minipage}
\hfill
	\begin{minipage}{0.3\textwidth}
	   \includegraphics[width=0.45\textwidth]{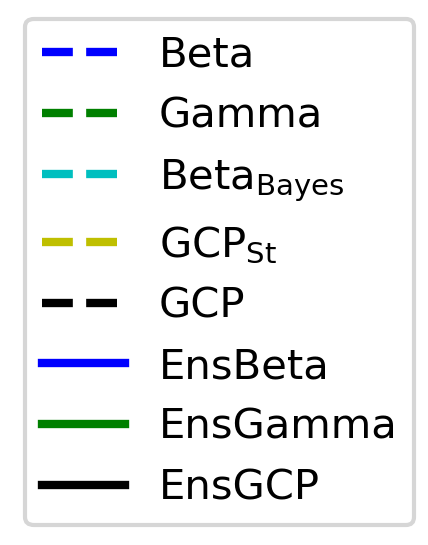}
	\end{minipage}

	\begin{minipage}{0.3\textwidth}
       \includegraphics[width=\textwidth]{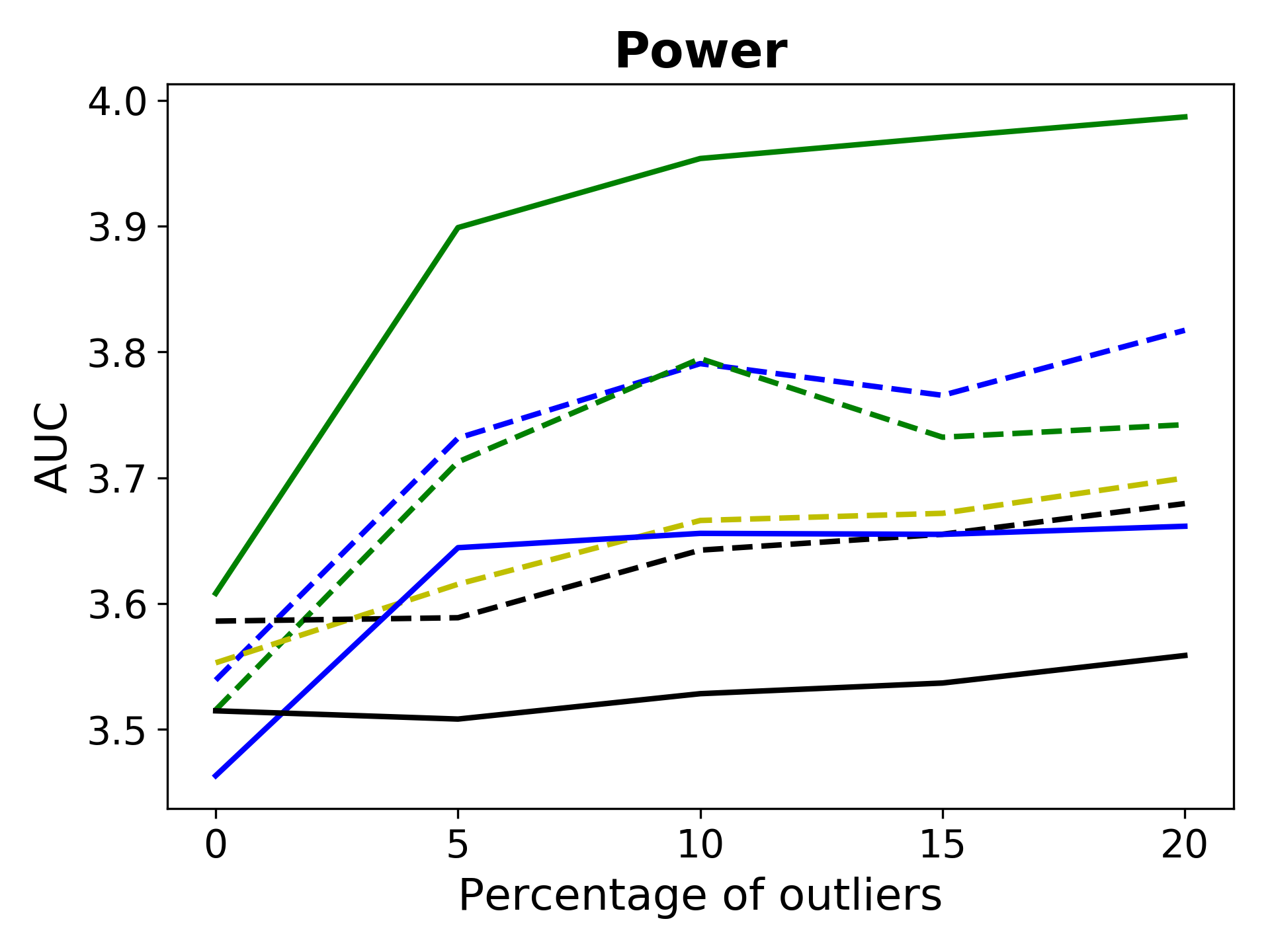}
	\end{minipage}
\hfill
	\begin{minipage}{0.3\textwidth}
       \includegraphics[width=\textwidth]{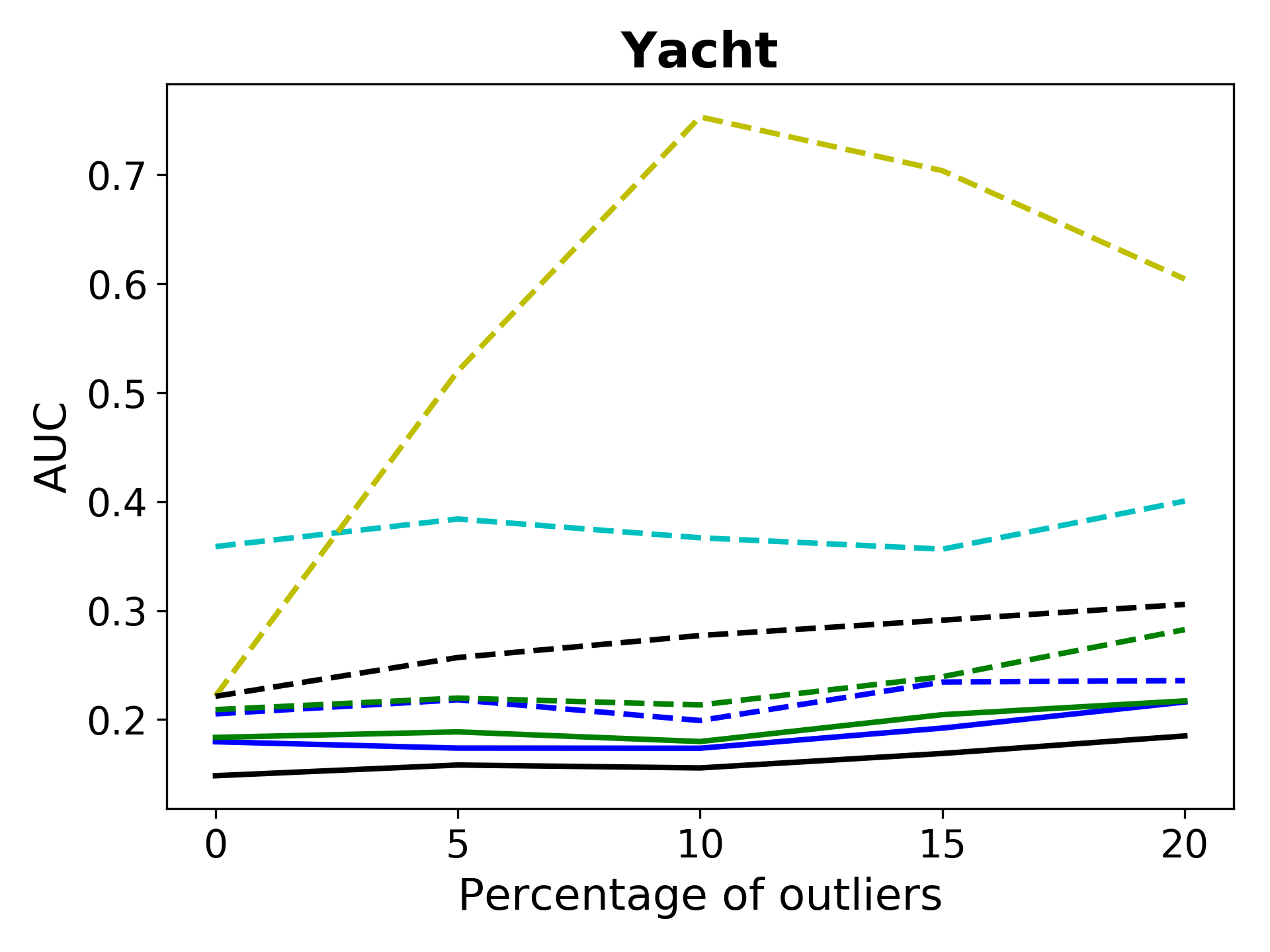}
	\end{minipage}
\hfill
	\begin{minipage}{0.3\textwidth}
       \includegraphics[width=\textwidth]{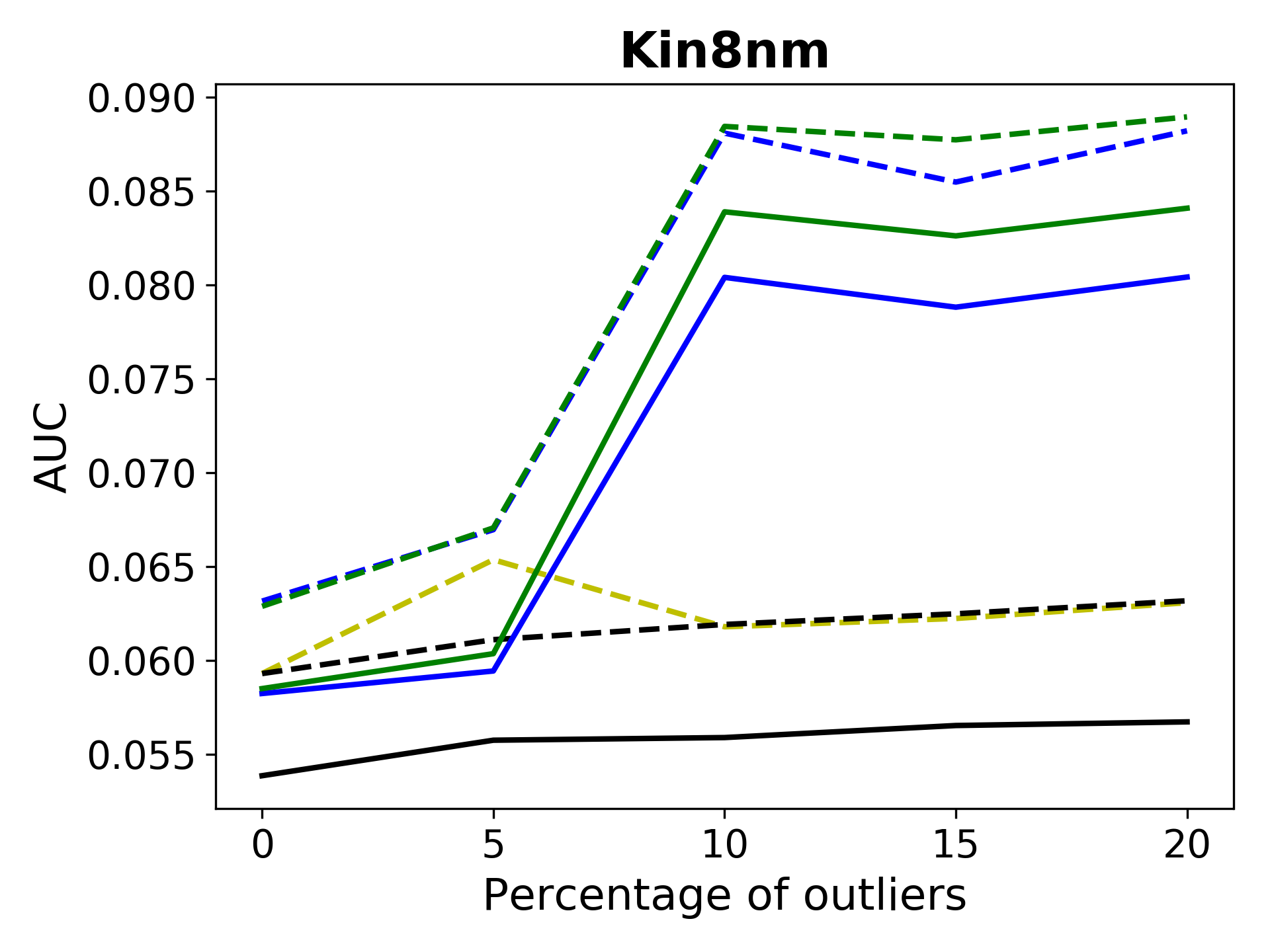}
	\end{minipage}
\caption{AUC scores of different methods from Table~\ref{tableRMSE} vs. the percentage of outliers.}\label{figAUCvsOutliers}
\end{center}
\vskip -0.2in
\end{figure*}

We compare the  following robust methods:\footnote{Our preliminary results with robust gradient descent in~\cite{HollandIkeda2018} and~\cite{Yin2018} were significantly worse than those obtained by the other methods, especially in case of input-dependent variance in the loss. Therefore we do not include them in Table~\ref{tableRMSE}. We did not implement the robust gradient estimation in~\cite{Prasad2018} because fine tuning its hyperparameters requires the knowledge of $\ep$, see Sec.~3.3 therein.}
\begin{enumerate}

\item {\bf Beta} and {\bf Gamma}: the methods in which one minimizes, respectively, the $\beta$- and  $\gamma$-divergences from the ground truth to the approximating normal distribution~\cite{Basu1998,Gosh2016,Fujisawa2008};

\item {\bf Beta$_{\rm Bayes}$}: the robust Bayesian method based on the $\beta$-divergence\footnote{We performed a grid search for $\beta$ and the (input-independent) standard deviation of the likelihood. By varying these two parameters, one obtains the same set of loss functions as by varying $\gamma$ and the standard deviation in the robust Bayesian method in~\cite{Futami2017} based on the $\gamma$-divergence. Therefore, we do not include the latter method as a separate one in our comparison list.}~\cite{Futami2017};

\item {\bf GCP$_{\rm St}$}: the GCP with the Student's t variance $\Vs$, \url{https://github.com/hstuk/GCP};

\item {\bf GCP}: the GCP with the prognostic variance $\Vp$, \url{https://github.com/hstuk/GCP};

\item {\bf EnsBeta}, {\bf EnsGamma}, {\bf EnsGCP}: ensembles of 5 Beta, Gamma, and GCP respectively.
\end{enumerate}

Note that the  Beta, Gamma and the GCP-based methods estimate aleatoric uncertainty since they learn the variance of labels conditioned on the input $x$, while the Bayesian method  Beta$_{\rm Bayes}$  estimates epistemic uncertainty since the variance of the likelihood is treated as a hyperparameter, while the predictive variance is $x$-dependent only due to randomness in the weights. The ensemble methods are supposed to learn both aleatoric and epistemic uncertainty, and their overall variance is computed as the variance of the Gaussian mixture distribution, cf.~\cite{Lakshminarayanan17}. Architectures and hyperparameters for all methods are given in the supplement.

\subsection{Synthetic data set}\label{subsecSynthetic}

We generate a synthetic data set containing 5\% of outliers. To do so, we choose the set $X$ consisting of 400 points uniformly distributed on the interval $(-1,1)$. For each $x\in X$, with probability 0.95 we sample $y$ from the normal distribution with mean $\sin(3x)$ and standard deviation $0.5\cos^4 x$, and with probability 0.05 we sample $y$ from a uniform distribution on the interval $(-4,16)$. Figure~\ref{figSyntheticDataWithOutliers} shows the data and the fits for different methods.
Even though the means are accurately predicted by most robust methods, the GCP learns the variance best. Furthermore, the output $\alpha$ of the GCP network provides additional information, namely, small values of $\alpha$ indicate that the corresponding samples belong to a (less trust-worthy) region in which the training set contained outliers.

\subsection{Real world data sets}\label{subsecRealWorld}

\textbf{Data sets.}
We analyze the following publicly available data sets: Boston House Prices~\cite{Harrison78} ($506$ samples, 13 features), Concrete Compressive Strength~\cite{ICheng98} ($1030$ samples, 8 features), Combined Cycle Power Plant~\cite{Tufekci14,Kaya12} ($9568$ samples, 4 features), Yacht Hydrodynamics~\cite{Gerritsma81,Ortigosa07} (308 samples, 6 features), and Kinematics of an 8 Link Robot Arm Kin8Nm\footnote{\url{http://mldata.org/repository/data/viewslug/regression-datasets-kin8nm/}} (8192 samples, 8 feature).
Each data set is randomly split into train-test subsets with 95\% of samples in the training subset. For each training set, we randomly choose $\lambda$\% of samples and replace them by outliers. The outliers are sampled from the Gaussian distribution with the mean equal to the mean over all the targets in the original training set and standard deviation equal to ten times the standard deviation over the targets in the original training set. All the results reported below are the respective averages over 50 cross-validations.

\textbf{Measures.} We use two measures of the quality of the fit.
1. The overall root mean squared error ({\em RMSE}).
2. The area under the following curve ({\em AUC}), measuring the trade-off between properly learning the mean and the variance. Assume the test set contains $N$ samples. We order them with respect to their predicted variance. For each $n=0,\dots,N-1$, we remove $n$  samples with the highest variance and calculate the RMSE for the remaining $N-n$ samples (with the lowest variance). We denote it by ${\rm RMSE}(n)$ and plot it versus~$n$ as a continuous piecewise linear curve. The second measure is the area under this curve normalized by $N-1$:
      $
      {\rm AUC} := \frac{1}{N-1}\sum\limits_{n=0}^{N-2}\frac{{\rm RMSE}(n)+{\rm RMSE}(n+1)}{2}.
      $

\textbf{Results.}
Table~\ref{tableRMSE} presents\footnote{Symbol $*$ indicates that we were not able to fine tune the parameters of the Beta$_{\rm Bayes}$ to obtain reasonable predictions for Power and Kin8nm data sets. Note that the authors in~\cite{Futami2017} used a protocol for fitting Beta$_{\rm Bayes}$ different from ours. Unlike us, they first normalized the noncontaminated training set and then added outliers.} RMSE and AUC scores for the outliers' percentage $\lambda=0,5,10,15,20$, respectively. In each column, we mark a method in bold if it is significantly (due to the two-tailed paired difference test with $p=0.05$) better or indistinguishable from {\em all} the other methods.  We see that the GCP significantly improves AUC scores of the GCP$_{\rm St}$ in the presence of outliers. Furthermore, EnsGCP yields the best AUC among all the methods for all $\lambda$, see also Fig.~\ref{figAUCvsOutliers}. Thus, it provides the best trade-off between properly learning the mean and the variance. Its RMSE score is competitive or superior to the other methods. Moreover, after removing a small number of samples for which EnsGCP predicts a high variance, its RMSE for the remaining samples becomes significantly better than the respective RMSE of the other methods, see the curves ${\rm RMSE}(n)$ in Fig.~5 in the supplement.

\begin{table*}[t]
\caption{RMSE and AUC scores for 0\%, 5\%, 10\%, 15\%, and 20\% of outliers.}\label{tableRMSE}
\begin{center}
\begin{small}
\begin{sc}
\resizebox{\textwidth}{!}{%
\begin{tabular}{lcc}
              &         {\bf Boston } &                    \\
\toprule
{Outliers: 0\%} &                  RMSE &                  AUC \\
\midrule
Beta               &  {\bf  3.59$\pm$1.51} &        2.14$\pm$0.49 \\
Gamma              &  {\bf  3.64$\pm$1.52} &        2.21$\pm$0.55 \\
Beta$_{\rm Bayes}$ &  {\bf  3.69$\pm$1.52} &        2.53$\pm$0.79 \\
GCP$_{\rm St}$     &  {\bf  3.62$\pm$1.60} &        1.92$\pm$0.42 \\
GCP                &  {\bf  3.62$\pm$1.60} &        1.91$\pm$0.41 \\
\midrule
EnsBeta            &         3.71$\pm$1.60 &        2.18$\pm$0.58 \\
EnsGamma           &         3.75$\pm$1.65 &        2.35$\pm$0.67 \\
EnsGCP             &  {\bf  3.67$\pm$1.61} &  {\bf 1.73$\pm$0.42} \\
\bottomrule
\toprule
{Outliers: 5\%} &                  RMSE &                  AUC \\
\midrule
Beta               &  {\bf  3.42$\pm$1.37} &        2.19$\pm$0.51 \\
Gamma              &         3.54$\pm$1.46 &        2.22$\pm$0.51 \\
Beta$_{\rm Bayes}$ &         3.76$\pm$1.56 &        2.58$\pm$0.81 \\
GCP$_{\rm St}$     &         3.57$\pm$1.47 &        2.55$\pm$1.11 \\
GCP                &         3.57$\pm$1.47 &        2.05$\pm$0.48 \\
\midrule
EnsBeta            &  {\bf  3.53$\pm$1.48} &        2.19$\pm$0.51 \\
EnsGamma           &         3.59$\pm$1.54 &        2.24$\pm$0.57 \\
EnsGCP             &  {\bf  3.61$\pm$1.52} &  {\bf 1.85$\pm$0.47} \\
\bottomrule
\toprule
{Outliers: 10\%} &                  RMSE &                  AUC \\
\midrule
Beta               &  {\bf  3.31$\pm$1.26} &        2.21$\pm$0.49 \\
Gamma              &         3.49$\pm$1.42 &        2.28$\pm$0.52 \\
Beta$_{\rm Bayes}$ &  {\bf  3.79$\pm$1.63} &        2.73$\pm$1.06 \\
GCP$_{\rm St}$     &         3.63$\pm$1.52 &        2.54$\pm$1.08 \\
GCP                &         3.63$\pm$1.52 &  {\bf 2.10$\pm$0.52} \\
\midrule
EnsBeta            &         3.49$\pm$1.46 &        2.18$\pm$0.53 \\
EnsGamma           &         3.55$\pm$1.52 &        2.22$\pm$0.52 \\
EnsGCP             &  {\bf  3.66$\pm$1.52} &  {\bf 1.90$\pm$0.52} \\
\bottomrule
\toprule
{Outliers: 15\%} &                  RMSE &                  AUC \\
\midrule
Beta               &  {\bf  3.32$\pm$1.24} &        2.33$\pm$0.51 \\
Gamma              &  {\bf  3.42$\pm$1.31} &  {\bf 2.21$\pm$0.53} \\
Beta$_{\rm Bayes}$ &         3.84$\pm$1.62 &        2.64$\pm$0.94 \\
GCP$_{\rm St}$     &         3.57$\pm$1.42 &  {\bf 2.33$\pm$0.99} \\
GCP                &         3.57$\pm$1.42 &  {\bf 2.18$\pm$0.67} \\
\midrule
EnsBeta            &  {\bf  3.45$\pm$1.42} &  {\bf 2.18$\pm$0.47} \\
EnsGamma           &         3.51$\pm$1.44 &  {\bf 2.19$\pm$0.47} \\
EnsGCP             &  {\bf  3.70$\pm$1.51} &  {\bf 2.03$\pm$0.64} \\
\bottomrule
\toprule
{Outliers: 20\%} &                  RMSE &                  AUC \\
\midrule
Beta               &  {\bf  3.49$\pm$1.32} &        2.69$\pm$0.72 \\
Gamma              &  {\bf  3.45$\pm$1.36} &        2.33$\pm$0.58 \\
Beta$_{\rm Bayes}$ &         3.84$\pm$1.58 &        2.60$\pm$1.06 \\
GCP$_{\rm St}$     &         3.68$\pm$1.52 &        2.49$\pm$1.02 \\
GCP                &         3.68$\pm$1.52 &        2.37$\pm$0.85 \\
\midrule
EnsBeta            &  {\bf  3.43$\pm$1.38} &        2.29$\pm$0.52 \\
EnsGamma           &         3.51$\pm$1.44 &  {\bf 2.23$\pm$0.50} \\
EnsGCP             &  {\bf  3.69$\pm$1.44} &  {\bf 2.06$\pm$0.54} \\
\bottomrule
\end{tabular}
\
\begin{tabular}{lcc}
      {\bf Concrete } &                    \\
\toprule
                 RMSE &                  AUC \\
\midrule
        6.58$\pm$1.09 &        4.09$\pm$0.86 \\
        6.54$\pm$1.11 &        4.12$\pm$0.91 \\
        6.06$\pm$0.97 &        5.60$\pm$1.07 \\
 {\bf  5.82$\pm$1.01} &        3.49$\pm$0.68 \\
 {\bf  5.82$\pm$1.01} &        3.49$\pm$0.72 \\
 \midrule
        6.38$\pm$1.05 &        3.91$\pm$0.75 \\
        6.23$\pm$1.02 &        3.89$\pm$0.66 \\
 {\bf  5.73$\pm$0.97} &  {\bf 3.37$\pm$0.71} \\
\bottomrule
\toprule
                 RMSE &                  AUC \\
\midrule
        6.25$\pm$1.07 &        4.05$\pm$0.80 \\
        6.20$\pm$0.99 &        4.06$\pm$0.78 \\
        6.01$\pm$0.96 &        5.40$\pm$1.14 \\
 {\bf  5.79$\pm$1.05} &        5.14$\pm$1.15 \\
 {\bf  5.79$\pm$1.05} &  {\bf 3.66$\pm$0.85} \\
 \midrule
        6.00$\pm$1.00 &        3.89$\pm$0.74 \\
        5.97$\pm$0.97 &        3.94$\pm$0.64 \\
 {\bf  5.68$\pm$1.01} &  {\bf 3.56$\pm$0.75} \\
\bottomrule
\toprule
                 RMSE &                  AUC \\
\midrule
        6.06$\pm$1.04 &        3.98$\pm$0.76 \\
        6.04$\pm$0.97 &        4.06$\pm$0.81 \\
        6.03$\pm$0.99 &        5.58$\pm$1.02 \\
 {\bf  5.71$\pm$0.96} &        4.50$\pm$1.11 \\
 {\bf  5.71$\pm$0.96} &        3.73$\pm$0.81 \\
 \midrule
        5.91$\pm$0.95 &        3.89$\pm$0.73 \\
        5.93$\pm$0.97 &        4.22$\pm$0.73 \\
 {\bf  5.69$\pm$1.01} &  {\bf 3.56$\pm$0.79} \\
\bottomrule
\toprule
                 RMSE &                  AUC \\
\midrule
        6.07$\pm$1.01 &        4.00$\pm$0.75 \\
        6.03$\pm$0.97 &        4.09$\pm$0.75 \\
        6.21$\pm$1.08 &        5.68$\pm$1.28 \\
 {\bf  5.76$\pm$1.06} &        3.80$\pm$0.80 \\
 {\bf  5.76$\pm$1.06} &        3.70$\pm$0.86 \\
 \midrule
        5.88$\pm$0.98 &        3.88$\pm$0.71 \\
        6.65$\pm$1.09 &        5.70$\pm$1.12 \\
 {\bf  5.70$\pm$1.03} &  {\bf 3.56$\pm$0.76} \\
\bottomrule
\toprule
                 RMSE &                  AUC \\
\midrule
        6.08$\pm$1.04 &        4.10$\pm$0.78 \\
        6.10$\pm$1.12 &        4.10$\pm$0.85 \\
        6.34$\pm$0.94 &        5.66$\pm$1.15 \\
 {\bf  5.83$\pm$0.98} &        3.88$\pm$0.87 \\
 {\bf  5.83$\pm$0.98} &        3.84$\pm$0.88 \\
 \midrule
        5.90$\pm$0.99 &        3.95$\pm$0.70 \\
        8.99$\pm$1.35 &        7.94$\pm$1.46 \\
 {\bf  5.75$\pm$1.05} &  {\bf 3.63$\pm$0.75} \\
\bottomrule
\end{tabular}
\
\begin{tabular}{lcc}
         {\bf Power } &                    \\
\toprule
                 RMSE &                  AUC \\
\midrule
        4.04$\pm$0.31 &        3.54$\pm$0.36 \\
        4.01$\pm$0.31 &        3.52$\pm$0.36 \\
      *               &              *       \\
        4.13$\pm$0.31 &        3.55$\pm$0.35 \\
        4.13$\pm$0.31 &        3.59$\pm$0.37 \\
        \midrule
 {\bf  3.97$\pm$0.31} &  {\bf 3.46$\pm$0.38} \\
        4.02$\pm$0.32 &        3.61$\pm$0.42 \\
        4.11$\pm$0.31 &        3.51$\pm$0.37 \\
\bottomrule
\toprule
                 RMSE &                  AUC \\
\midrule
        4.05$\pm$0.30 &        3.73$\pm$0.37 \\
        4.02$\pm$0.31 &        3.71$\pm$0.39 \\
      *               &              *       \\
        4.15$\pm$0.31 &        3.62$\pm$0.36 \\
        4.15$\pm$0.31 &        3.59$\pm$0.33 \\
        \midrule
 {\bf  3.96$\pm$0.31} &        3.64$\pm$0.39 \\
        4.03$\pm$0.33 &        3.90$\pm$0.59 \\
        4.12$\pm$0.31 &  {\bf 3.51$\pm$0.34} \\
\bottomrule
\toprule
                 RMSE &                  AUC \\
\midrule
        4.07$\pm$0.31 &        3.79$\pm$0.42 \\
        4.04$\pm$0.32 &        3.79$\pm$0.50 \\
      *               &              *       \\
        4.17$\pm$0.31 &        3.67$\pm$0.37 \\
        4.17$\pm$0.31 &        3.64$\pm$0.35 \\
        \midrule
 {\bf  3.96$\pm$0.31} &        3.66$\pm$0.39 \\
        4.03$\pm$0.33 &        3.95$\pm$0.66 \\
        4.13$\pm$0.31 &  {\bf 3.53$\pm$0.35} \\
\bottomrule
\toprule
                 RMSE &                  AUC \\
\midrule
        4.09$\pm$0.31 &        3.77$\pm$0.36 \\
        4.05$\pm$0.31 &        3.73$\pm$0.37 \\
      *               &              *       \\
        4.19$\pm$0.30 &        3.67$\pm$0.28 \\
        4.19$\pm$0.30 &        3.66$\pm$0.28 \\
        \midrule
 {\bf  3.98$\pm$0.31} &        3.66$\pm$0.36 \\
        4.03$\pm$0.32 &        3.97$\pm$0.65 \\
        4.14$\pm$0.31 &  {\bf 3.54$\pm$0.33} \\
\bottomrule
\toprule
                 RMSE &                  AUC \\
\midrule
        4.14$\pm$0.30 &        3.82$\pm$0.36 \\
        4.08$\pm$0.31 &        3.74$\pm$0.32 \\
      *               &              *       \\
        4.22$\pm$0.31 &        3.70$\pm$0.33 \\
        4.22$\pm$0.31 &        3.68$\pm$0.32 \\
        \midrule
 {\bf  3.99$\pm$0.32} &        3.66$\pm$0.34 \\
        4.04$\pm$0.33 &        3.99$\pm$0.63 \\
        4.14$\pm$0.31 &  {\bf 3.56$\pm$0.33} \\
\bottomrule
\end{tabular}
\
\begin{tabular}{lcc}
         {\bf Yacht } &                    \\
\toprule
                 RMSE &                  AUC \\
\midrule
        0.99$\pm$0.48 &        0.21$\pm$0.08 \\
        0.94$\pm$0.49 &        0.21$\pm$0.07 \\
 {\bf  0.78$\pm$0.34} &        0.36$\pm$0.16 \\
        1.07$\pm$0.58 &        0.22$\pm$0.09 \\
        1.07$\pm$0.58 &        0.22$\pm$0.09 \\
        \midrule
        0.89$\pm$0.45 &        0.18$\pm$0.07 \\
        0.93$\pm$0.49 &        0.18$\pm$0.07 \\
 {\bf  0.71$\pm$0.40} &  {\bf 0.15$\pm$0.07} \\
\bottomrule
\toprule
                 RMSE &                  AUC \\
\midrule
        0.91$\pm$0.49 &        0.22$\pm$0.08 \\
        0.91$\pm$0.43 &        0.22$\pm$0.08 \\
        0.78$\pm$0.32 &        0.38$\pm$0.13 \\
        0.99$\pm$0.55 &        0.52$\pm$0.29 \\
        0.99$\pm$0.55 &        0.26$\pm$0.10 \\
        \midrule
        0.78$\pm$0.41 &        0.17$\pm$0.06 \\
        0.82$\pm$0.44 &        0.19$\pm$0.08 \\
 {\bf  0.58$\pm$0.31} &  {\bf 0.16$\pm$0.05} \\
\bottomrule
\toprule
                 RMSE &                  AUC \\
\midrule
        0.84$\pm$0.48 &        0.20$\pm$0.06 \\
        0.81$\pm$0.48 &        0.21$\pm$0.08 \\
        0.80$\pm$0.34 &        0.37$\pm$0.12 \\
        1.00$\pm$0.52 &        0.75$\pm$0.52 \\
        1.00$\pm$0.52 &        0.28$\pm$0.13 \\
        \midrule
        0.70$\pm$0.38 &        0.17$\pm$0.06 \\
        0.77$\pm$0.43 &        0.18$\pm$0.07 \\
 {\bf  0.58$\pm$0.31} &  {\bf 0.16$\pm$0.05} \\
\bottomrule
\toprule
                 RMSE &                  AUC \\
\midrule
        0.87$\pm$0.38 &        0.23$\pm$0.07 \\
        0.95$\pm$0.44 &        0.24$\pm$0.07 \\
        0.80$\pm$0.35 &        0.36$\pm$0.11 \\
        1.09$\pm$0.46 &        0.70$\pm$0.51 \\
        1.09$\pm$0.46 &        0.29$\pm$0.10 \\
        \midrule
        0.75$\pm$0.37 &        0.19$\pm$0.06 \\
        0.87$\pm$0.43 &        0.20$\pm$0.07 \\
 {\bf  0.58$\pm$0.28} &  {\bf 0.17$\pm$0.06} \\
\bottomrule
\toprule
                 RMSE &                  AUC \\
\midrule
        0.86$\pm$0.40 &        0.24$\pm$0.08 \\
        1.03$\pm$0.48 &        0.28$\pm$0.10 \\
        0.91$\pm$0.38 &        0.40$\pm$0.12 \\
        1.03$\pm$0.56 &        0.60$\pm$0.71 \\
        1.03$\pm$0.56 &        0.31$\pm$0.11 \\
        \midrule
        0.77$\pm$0.40 &        0.22$\pm$0.08 \\
        0.91$\pm$0.44 &        0.22$\pm$0.07 \\
 {\bf  0.61$\pm$0.30} &  {\bf 0.18$\pm$0.07} \\
\bottomrule
\end{tabular}
\
\begin{tabular}{lcc}
       {\bf Kin8nm } &                     \\
\toprule
                RMSE &                   AUC \\
\midrule
       0.11$\pm$0.02 &         0.06$\pm$0.00 \\
       0.11$\pm$0.02 &         0.06$\pm$0.00 \\
                   * &                     * \\
       0.09$\pm$0.01 &         0.06$\pm$0.00 \\
       0.09$\pm$0.01 &         0.06$\pm$0.00 \\
       \midrule
       0.10$\pm$0.01 &         0.06$\pm$0.00 \\
       0.11$\pm$0.01 &         0.06$\pm$0.00 \\
 {\bf 0.08$\pm$0.00} &  {\bf  0.05$\pm$0.00} \\
\bottomrule
\toprule
                RMSE &                   AUC \\
\midrule
       0.09$\pm$0.01 &         0.07$\pm$0.00 \\
       0.09$\pm$0.01 &         0.07$\pm$0.00 \\
                   * &                     * \\
       0.09$\pm$0.01 &         0.07$\pm$0.01 \\
       0.09$\pm$0.01 &         0.06$\pm$0.00 \\
       \midrule
       0.08$\pm$0.01 &         0.06$\pm$0.00 \\
       0.09$\pm$0.01 &         0.06$\pm$0.00 \\
 {\bf 0.08$\pm$0.00} &  {\bf  0.06$\pm$0.00} \\
\bottomrule
\toprule
                RMSE &                   AUC \\
\midrule
       0.09$\pm$0.01 &         0.09$\pm$0.01 \\
       0.09$\pm$0.01 &         0.09$\pm$0.01 \\
                   * &                     * \\
       0.09$\pm$0.01 &         0.06$\pm$0.00 \\
       0.09$\pm$0.01 &         0.06$\pm$0.00 \\
       \midrule
       0.08$\pm$0.01 &         0.08$\pm$0.01 \\
       0.09$\pm$0.01 &         0.08$\pm$0.01 \\
 {\bf 0.08$\pm$0.00} &  {\bf  0.06$\pm$0.00} \\
\bottomrule
\toprule
                RMSE &                   AUC \\
\midrule
       0.09$\pm$0.01 &         0.09$\pm$0.01 \\
       0.09$\pm$0.01 &         0.09$\pm$0.01 \\
                   * &                     * \\
       0.09$\pm$0.01 &         0.06$\pm$0.00 \\
       0.09$\pm$0.01 &         0.06$\pm$0.00 \\
       \midrule
       0.08$\pm$0.01 &         0.08$\pm$0.01 \\
       0.09$\pm$0.01 &         0.08$\pm$0.01 \\
 {\bf 0.08$\pm$0.00} &  {\bf  0.06$\pm$0.00} \\
\bottomrule
\toprule
                RMSE &                   AUC \\
\midrule
       0.09$\pm$0.01 &         0.09$\pm$0.01 \\
       0.09$\pm$0.01 &         0.09$\pm$0.01 \\
                   * &                     * \\
       0.09$\pm$0.01 &         0.06$\pm$0.00 \\
       0.09$\pm$0.01 &         0.06$\pm$0.00 \\
       \midrule
       0.08$\pm$0.01 &         0.08$\pm$0.01 \\
       0.08$\pm$0.01 &         0.08$\pm$0.01 \\
 {\bf 0.08$\pm$0.00} &  {\bf  0.06$\pm$0.00} \\
\bottomrule
\end{tabular}

}
\end{sc}
\end{small}
\end{center}
\vskip -0.1in
\end{table*}

\section{Conclusion}

We analyzed the minima of the energy surfaces of the GCP networks encoding the priors of latent variable models. Under the assumption of Huber's $\ep$-contamination of the Gaussian ground truth distribution $\pg(y|x)$, we obtained formulas for prognostic mean $\mp(x)$ and variance $\Vp(x)$ in terms of the outputs of the GCP networks, yielding errors for the ground truth mean $\mg(x)$ and variance $\Vg(x)$ of order $O(e^{-c/\ep})$ and $O(\ep)$ respectively. 


The GCP networks can be trained with standard optimizers (such as Adam, RMSProp, etc.) and do not require fine tuning additional hyperparameters. Experiments with synthetic and real world data with outliers showed their superiority over several other state-of-art robust methods based on neural networks.

{\small

\bibliography{bibl}
\bibliographystyle{icml2019}
}

\appendix

\section{Algorithm for fitting a GCP network and predicting the mean and variance of the ground truth distribution}

In this section, we present a practical algorithm for defining a loss of a GCP network, fitting it, and predicting the mean and variance of the ground truth distribution in a robust way. The code is available at~\url{https://github.com/hstuk/GCP}.

Given an input $x\in \mathbb R^d$ and a vector of weights $w$, we denote the $4$-dimensional output of the GCP network by $m(w,x),\alpha(w,x),\beta(w,x),\nu(w,x)$. The outputs can share the weights or have independent weights, in which case $w=(w_m,w_\alpha,w_\beta,w_\nu)$.
For each labeled sample $(x,y)$ with $x\in\mathbb R^d$, $y\in\mathbb R$, we define a loss $L(w,x,y)$ according to Algorithm~\ref{algLossKL} or Algorithm~\ref{algLossStudent}. According to~\cite[Lemma~2.1]{GurHannesGCP}, these two algorithms yield the same loss up to an additive constant not depending on $w$.
\begin{algorithm}[t]
   \caption{Loss $L(w,x,y)$ of a GCP network via the KL-divergence}
   \label{algLossKL}
\begin{algorithmic}
   \STATE {\bfseries Input:} Vector of weights $w$ and a labeled sample $(x,y)$
   \STATE Fix current weights: $w_{\rm fix}  \leftarrow w$
   \STATE Fix current prior parameters:
   $$
   (m_{\rm fix},\alpha_{\rm fix},\beta_{\rm fix},\nu_{\rm fix}) \leftarrow (m(w_{\rm fix},x), \alpha(w_{\rm fix},x),\beta(w_{\rm fix},x),\nu(w_{\rm fix},x))
   $$
   \STATE Compute the parameters of the posterior:
   $$
   (m',\alpha',\beta',\nu') \leftarrow \left(\dfrac{\nu_{\rm fix} m_{\rm fix} +  y }{\nu_{\rm fix} + 1},
 \alpha_{\rm fix}+\frac{1}{2}, \beta_{\rm fix} +   \frac{\nu_{\rm fix}}{\nu_{\rm fix}+1}\frac{( y - m_{\rm fix})^2}{2}, \nu_{\rm fix} + 1\right).
   $$
   \STATE Compute the KL-divergence:
   $$
   \begin{aligned}
& K(w,x,y) \leftarrow   \frac{\alpha'(m(w,x)-m')^2\nu}{2\beta'} +  \frac{\nu(w,x)}{2\nu'} - \frac{1}{2}\ln\frac{\nu(w,x)}{\nu'}-\frac{1}{2}\\
& - \alpha(w,x)\ln\frac{\beta(w,x)}{\beta'} + \ln\frac{\Gamma(\alpha(w,x))}{\Gamma(\alpha')} - (\alpha(w,x)-\alpha')\Psi(\alpha') + \frac{\alpha'(\beta(w,x)-\beta')}{\beta'},
\end{aligned}
$$
   \STATE {\bfseries Return:} loss function $L(w,x,y)=-K(w,x,y)$
\end{algorithmic}
\end{algorithm}

\begin{algorithm}[t]
   \caption{Loss $L(w,x,y)$ of a GCP network via the log-likelihood of Student's t-distribution}
   \label{algLossStudent}
\begin{algorithmic}
   \STATE {\bfseries Input:} Vector of weights $w$ and a labeled sample $(x,y)$
   \STATE Define the parameters of Student's t-distribution:
   $$
   (\tilde{\nu}(w,x),\tilde{\sigma}(w,x)) \leftarrow \left(2\alpha(w,x), \frac{\beta(w,x)(\nu(w,x)+1)}{\nu(w,x)\alpha(w,x)}\right)
   $$
   \STATE Compute the likelihood of Student's t-distribution:
   $$
   t(w,x,y) \leftarrow \frac{\Gamma\left(\frac{\tilde\nu(w,x)+1}{2}\right)}{\Gamma\left(\frac{\tilde\nu(w,x)}{2}\right)\sqrt{\pi\tilde\nu(w,x)}\tilde\sigma(w,x)}
   \left(1+\frac{1}{\tilde\nu(w,x)}\left(\frac{y-m(w,x)}{\tilde\sigma(w,x)}\right)^2\right)^{-\frac{\tilde\nu(w,x)+1}{2}}
   $$
   \STATE {\bfseries Return:} loss function $L(w,x,y)=-\ln t(w,x,y)$
\end{algorithmic}
\end{algorithm}

\begin{figure}[b]
\begin{center}
\centerline{\includegraphics[width=0.6\linewidth]{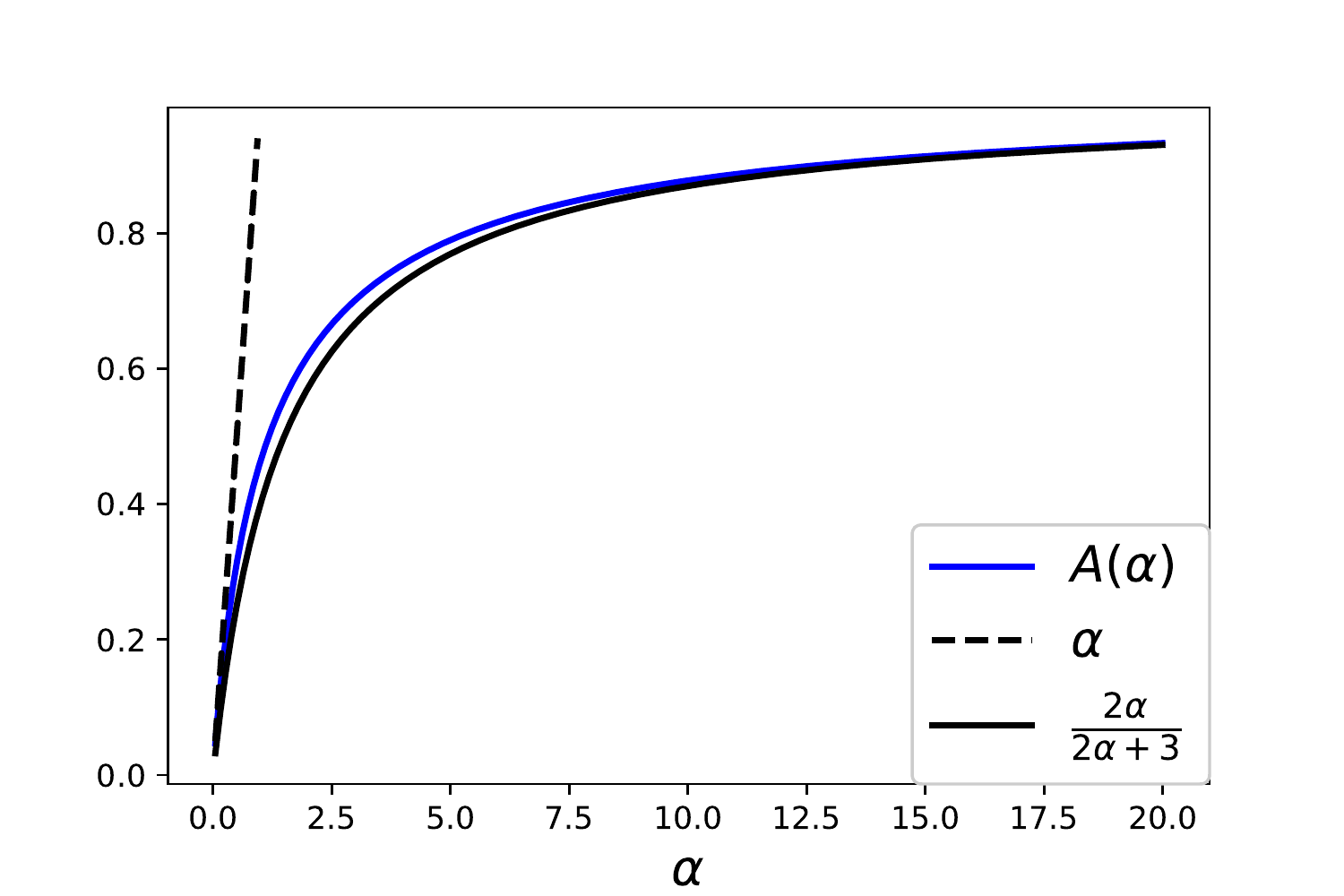}}
\caption{The graph of $A(\alpha)$ entering Eq.~(7).}
\label{figA}
\end{center}
\end{figure}

Given the loss $L(w,x,y)$ defined in Algorithm~\ref{algLossKL} or~\ref{algLossStudent} and a training set $(X,Y)$, we fit the GCP network by minimizing
$$\sum\limits_{(x,y)\in (X,Y)} L(w,x,y),$$
using any standard optimizer (e.g., Adam, RMSProp, etc.). Once the GCP network is fitted, we predict the mean and variance of the ground truth distribution $\pg(y|x)$ as follows (see Eq.~(6)):
$$
\mp(x):=m(w,x),\qquad \Vp(x) := \dfrac{\beta(w,x)(\nu(w,x)+1)}{\nu(w,x)(\alpha(w,x)-A(\alpha(w,x)))},
$$
where $A(\alpha)$ is defined as a unique root of Eq.~(7). The function $A(\alpha)$ can be precalculated in advance or, due to~\cite{GurHannesGCP}, approximated by
$$
A(\alpha)\approx \frac{2\alpha}{2\alpha+3},
$$
see Fig.~\ref{figA}.

\begin{remark}
  The fitted GCP network minimizes the log-likelihood of Student's t-distribution $p(y|x,m(w,x),\tilde\nu(w,x),\tilde\sigma(w,x))$, see Algorithm~\ref{algLossStudent}. One can rewrite the above prognostic variance $\Vp(x)$ in terms of $\tilde\nu(w,x),\tilde\sigma(w,x)$, namely
  $$
  \Vp(x) := \dfrac{\tilde\sigma(w,x)\tilde\nu(w,x)}{\tilde\nu(w,x)-2A(\tilde\nu(w,x)/2))}.
  $$
  This approach would reduce the $4$-dimensional output of the GCP network to the $3$-dimensional output directly encoding the parameters of Student's t distribution. However, the resulting dynamics of the weights $w$ and the induced dynamics of $m(w,x),\tilde\sigma(w,x),\tilde\sigma(w,x)$ (a counterpart for dynamical system~(12)--(14)) is an open question, which is a direction for future research.
\end{remark}

\section{Proof of Theorem~$3.1$}\label{secSmallEpEquilibriumProof}

We assume throughout the proof that $\po(y)$ is continuously differentiable, its sixth central moment exists, and there is $C>0$ such
\begin{equation}\label{eqpoPrime}
  |\po'(y)|\le \frac{C}{|y|^2}\quad \text{for all } |y|\ge 1,
\end{equation}
and
\begin{equation}\label{eqpoIntegral}
  \int_{M}^{\infty}\po(y)\le \frac{C}{M^2},\quad \int_{-\infty}^{-M}\po(y)\le \frac{C}{M^2}
\end{equation}
for all  $M\ge 1$.

1. Without loss of generality, assume that
\begin{equation}\label{eqmg0}
  \mg=0.
\end{equation}
First, we show that system~(12)--(14) has at least one equilibrium $m_\ep,\alpha_\ep,\beta_\ep,\nu_\ep$. To do so, it suffices to prove that the system of equations
\begin{align}
F(m,\sigma,\ep)& = \int
\frac{z}{2\sigma  + z^2}\pc(y)\,dy=0,\label{eqDKmPureZero}\\
{}
G(m,\alpha,\sigma,\ep) & =  \int\ln\left(1+ \frac{z^2}{2\sigma}\right)\pc(y)\,dy + \Delta\Psi(\alpha) =0,\label{eqDKalphaPureZero}\\
{}
H(m,\alpha,\sigma,\ep) & = \int\frac{\alpha z^2 - \sigma}{2\sigma + z^2}\pc(y)\,dy=0\label{eqDKsigmaPureZero}
\end{align}
(where the integrals are take over $\bbR$, $z=y-m$, $\pc(y)$ is defined in~(1), and
$\Delta\Psi(\alpha):=\Psi(\alpha)-\Psi(\alpha+1/2)$) has a root $m_\ep,\alpha_\ep,\sigma_\ep$.

First, we solve Eq.~\eqref{eqDKsigmaPureZero} with respect to $\alpha$.
Setting $\delta = 1/(2\sigma)$, we have for $ z^2 < 2 \sigma $:
\[
\frac{1}{2\sigma + z^2} = \delta \left[1 - \delta z^2   +\delta^2 z^4 - \delta^3 z^6 + O(\delta^4 z^8) \right]
\]
Hence, additionally using the decay of $\pc(y)$ at infinity to estimate the integral for $z^2>2\sigma$, we obtain
\begin{align*}
& H(m,\alpha,\sigma,\ep) = \int\limits_{\{z^2<2\sigma\}}  \left(\alpha z^2 - \frac{1}{2 \delta}\right)\delta\\
 &\left[1 - \delta z^2   +\delta^2 z^4 - \delta^3 z^6 + O(\delta^4 z^8) \right] p_c(y) dy \\
 & + f_0(m,\delta,\ep) \\
&= \alpha \left( \delta  V(m,\ep) - \delta^2  K(m,\ep) + \delta ^3  N(m,\ep) + f_1(m,\delta,\ep) \right) \\
& - \frac{1}{2} + \frac{\delta V(m,\ep)}{2} - \delta^2 \frac{K(m,\ep)}{2}  + \delta^3 \frac{N(m,\ep)}{2} \\
&  + f_2(m,\delta,\ep).
\end{align*}
Here
$$
\begin{aligned}
& V(m,\ep) = \int (y-m)^2\pc(y)\,dy,\\
& K(m,\ep) = \int (y-m)^4\pc(y)\,dy,\\
& N(m,\ep) = \int (y-m)^6\pc(y)\,dy,
\end{aligned}
$$
the functions $f_j(m,\delta,\ep)$ for $j=0,1,2$ (and $j=3,4$ below) are smooth for $\ep\in[0,1]$ and $m,\delta$ in a neighborhood of the origin, and their partial derivatives with respect to $m$ and $\ep$ are $O(\delta^4)$ as $\delta\to 0$ uniformly for $\ep\in[0,1]$ and $m$ in a neighborhood of the origin.

Solving $H(m,\alpha,\sigma,\ep)=0$ for $\alpha$ yields
\begin{equation}\label{eqAlphaFromH}
\alpha =\frac{1}{2\delta} \frac{1-\delta V + \delta^2 K - \delta^3 N - f_2}{V-\delta K + \delta^2 N +\delta^{-1}f_1},
\end{equation}
where, for brevity, we omitted the dependence of the functions on their arguments.

Using the Taylor formula for the logarithm and the asymptotic expansion of $\Psi(\alpha)$, we have
\begin{equation}\label{eqGExpanded}
\begin{aligned}
G(m,\alpha,\sigma,\ep) & = \delta V - \frac{\delta^2 K}{2} + \frac{\delta^3 N}{3} + f_3(m,\delta,\ep)\\
 & - \frac{1}{2 \alpha} - \frac{1}{8 \alpha^2} + O(\alpha^{-4}).
\end{aligned}
\end{equation}
Plugging in $ \alpha $ given by~\eqref{eqAlphaFromH} into~\eqref{eqGExpanded} and dividing by $\delta^2$, we see that, for $\delta>0$, system~\eqref{eqDKmPureZero}--\eqref{eqDKsigmaPureZero} is  equivalent to
\begin{align}
F_1(m,\delta,\ep)& := \int
\frac{z}{1  + \delta z^2}\pc(y)\,dy=0,\label{eqDKmPureZeroEquiv}\\
{}
F_2(m,\delta,\ep) & := \frac{K - 3 V^2}{2} - \frac{6V^3-9KV+2N}{3}  \delta\\
& + \delta^{-2} f_4(m,\delta,\ep) =0\label{eqDKsigmaPureZeroEquiv}
\end{align}
We solve system~\eqref{eqDKmPureZeroEquiv}--\eqref{eqDKsigmaPureZeroEquiv} with respect to $m,\delta$, using the implicit function theorem. Note that $F_1(0,0,0)=\int y\pg(y)\,dy=0$ due to~\eqref{eqmg0} and $F_2(0,0,0)=K(0,0,0) - 3 V^2(0,0,0)=0$ since $V(0,0,0)$ and $K(0,0,0)$ are the second and the fourth central moments of the Gaussian distribution $\pg(y)$. At $(m,\delta,\ep)=(0,0,0)$, we have
$$
\begin{aligned}
J  :=
\begin{bmatrix}
\partial_m F_1 & \partial_\delta F_1 \\
\partial_m F_2 & \partial_\delta F_2
\end{bmatrix}  =
\begin{bmatrix}
-1 & 0 \\
0 & -\frac{6V^3-9KV+2N}{3}\Big|_{m=0,\delta=0,\ep=0}
\end{bmatrix}  =
\begin{bmatrix}
-1 & 0 \\
0 & -3\Vg^3
\end{bmatrix}.
\end{aligned}
$$
The vector of $\ep$-derivatives at $(0,0,0)$ is
$$
\begin{aligned}
\begin{bmatrix}
\partial_\ep F_1\\
\partial_\ep F_2
\end{bmatrix}   =
\begin{bmatrix}
\mo\\
\frac{\partial_\ep K - 6 V \partial_\ep V}{2}\Big|_{m=0,\delta=0,\ep=0}
\end{bmatrix}   =
\begin{bmatrix}
\mo\\
\frac{-3\Vg^2 + \omegao^{(4)} - 6 \Vg(-\Vg + \omegao^{(2)})}{2}
\end{bmatrix}
 =
\begin{bmatrix}
\mo\\
\frac{\Cgo}{2}
\end{bmatrix}
\end{aligned}
$$
Hence, by the implicit function theorem, there exist $m_1,\delta_1,\ep_1>0$ such that for any $\ep\in[0,\ep_1]$, system~\eqref{eqDKmPureZeroEquiv}, \eqref{eqDKsigmaPureZeroEquiv} has a unique root $m_\ep,\delta_\ep$ in the set
\begin{equation}\label{eqNeighborhood1}
\{|m|<m_1,|\delta|<\delta_1\}.
\end{equation}
Moreover, $m_\ep,\delta_\ep$ are smooth functions of $\ep$ and
\begin{equation}\label{eqPartialEp}
\begin{bmatrix}
\partial_\ep m_\ep\\
\partial_\ep \delta_\ep
\end{bmatrix} =
-J^{-1}\begin{bmatrix}
\partial_\ep F_1\\
\partial_\ep F_2
\end{bmatrix} =
\begin{bmatrix}
\mo\\
\frac{\Cgo}{6\Vg^3}
\end{bmatrix}.
\end{equation}
In particular, \eqref{eqPartialEp} shows that $\delta_\ep>0$ and hence $\sigma_\ep>0$.  Combining~\eqref{eqPartialEp}  with~\eqref{eqAlphaFromH} proves asymptotics~(22). To prove asymptotics~(21), we substitute $m_\ep=\mo\ep + M\ep^2 + O(\ep^3)$ and $\delta_\ep = \frac{\Cgo}{6\Vg^3}\ep + O(\ep^2) $ into~\eqref{eqDKmPureZeroEquiv}. This yields
$$
\begin{aligned}
F_1(m_\ep,\delta_\ep,\ep) & = \left(M + \frac{\Cgo}{6\Vg^3} (\omegao^{(3)} -3\Vg\mo)\right)\ep^2 
 + O(\ep^3)=0,
\end{aligned}
$$
where $\omegao^{(3)}$ is the third moment about $0$ for the outliers distribution $\po(y)$.
Rewriting $\omegao^{(3)}$ via the central moments, we see that the constant $M$ equals the coefficient at $\ep^2$ in~(21).

2. It remains to show that system~(12)--(14) has no other equilibrium except for that found in part 1 of the proof. Assume, to the contrary, that there is a sequence $\ep_n\to 0$ and the respective sequence of solutions $(m_n,\alpha_n,\sigma_n)$ of system~\eqref{eqDKmPureZero}--\eqref{eqDKsigmaPureZero} that is different for each $\ep_n$ from those in part 1 of the proof.

First, we show that there exists $\tilde m$ (independent of $\sigma>0$ and $\ep\in[0,1]$) such that $|m_n|\le\tilde m$. Assume this is not true.
First consider the case where $\sigma_n$ is bounded. Let $m_n\to -\infty$ (the case $m_n\to\infty$) is treated similarly. We rewrite Eq.~\eqref{eqDKmPureZero} as follows:
\begin{equation}\label{eqI123}
I_1+I_2+I_3=0,
\end{equation}
where
$$
\begin{aligned}
I_1&:=\int_{-\infty}^{-1}
\frac{z}{2\sigma_n  + z^2}\pc(z+m_n)\,dz,\\
I_2&:=\int_{-1}^{1}
\frac{z}{2\sigma_n  + z^2}\pc(z+m_n)\,dz,\\
I_3&:=\int_{m_n+1}^{\infty}
\frac{y-m_n}{2\sigma_n  + (y-m_n)^2}\pc(y)\,dy.
\end{aligned}
$$
Using~\eqref{eqpoIntegral} and~\eqref{eqpoPrime}, we have
\begin{equation}\label{eqI1}
\begin{aligned}
  |I_1|&\le \int_{-\infty}^{-1} \frac{z^2}{2\sigma_n  + z^2} \pc(z+m_n)\,dz\\
  & \le
  \int_{-\infty}^{m_n-1} \pc(z)\,dz \le \frac{C_1}{m_n^2},
  \end{aligned}
\end{equation}
\begin{equation}\label{eqI2}
\begin{aligned}
  |I_2| &=  \int_{-1}^{1} \frac{z^2}{2\sigma_n  + z^2} \left|\frac{\pc(z+m_n)-\pc(m_n)}{z}\right|\,dz\\
   &\le 2\max\limits_{z\in[m_n-1,m_n+1]}|p'(z)|\le \frac{C_2}{m_n^2},
  \end{aligned}
\end{equation}
where $C_1,C_2>0$ do not depend on $n$. Further, we choose $M>0$ such that
$$
\int_{-M}^{M} \pc(y)\,dy \ge \frac{1}{2}
$$
for all $n>0$. Then, using the assumption that $\sigma_n$ is bounded, we have for all sufficiently large $n$
\begin{equation}\label{eqI3}
\begin{aligned}
 I_3 & \ge \int_{-M}^{M}
\frac{y-m_n}{2\sigma_n  + (y-m_n)^2}\pc(y)\,dy\\
& \ge \frac{C_3}{m_n}\int_{-M}^{M}\pc(y)\,dy\ge\frac{C_3}{2|m_n|},
  \end{aligned}
\end{equation}
where $C_3>0$ does not depend on $n$.
Relations~\eqref{eqI1}--\eqref{eqI3} contradict~\eqref{eqI123}.

%

Consider the case $\sigma_n\to\infty$. Then $\delta_n:=1/(2\sigma_n) \to 0$,  and we rewrite Eq.~\eqref{eqDKmPureZero} as follows:
$$
\tilde F(m_n,\delta_n,\ep_n) := \int
\frac{y-m_n}{1  + \delta_n (y-m_n)^2}\pc(y)\,dy=0,
$$
Then
$$
\begin{aligned}
0 & = \tilde F(m_n,\delta_n,\ep_n) - \tilde F(m_n,0,0) + \mc-m_n \\
& = g_1(\ep_n,\delta_n,m_n) + g_2(\ep_n,\delta_n,m_n)m_n + \mg - m_n,
\end{aligned}
$$
where $g_1(\delta,\ep,m),g_2(\delta,\ep,m)\to 0$ as $(\delta,\ep)\to 0$ uniformly with respect to $m\in\bbR$. This again contradicts the assumption $m_n\to\infty$. Thus, any root of Eq.~\eqref{eqDKmPureZero} indeed satisfies $|m|\le \tilde m$.

3. Further, we show that $\sigma_n$ is bounded away from $0$. Assume, to the contrary, that (possibly after passing to a subsequence) $\sigma_n\to 0$. Then, due to~\eqref{eqDKsigmaPureZero}, $\alpha_n\to 0$. Expressing $\alpha$ via $\sigma$ in~\eqref{eqDKsigmaPureZero} and using the fact that $m_n$ is bounded, we immediately see that $\alpha_n\le c_1\sqrt{\sigma_n}$ for all sufficiently large $n$, where $c_1>0$ does not depend on  $n$. On the other hand, \eqref{eqDKalphaPureZero} is equivalent to
$$
\int\ln(2\sigma_n + z^2) - \ln(2\sigma_n) - \frac{1}{\alpha_n} + O(1) = 0.
$$
Since $m_n$ is bounded, the latter equality yields $\alpha_n\ge c_2/\ln(\sigma^{-1})$ for all sufficiently large  $n$, where $c_2>0$ does not depend on $n$. This contradicts the first inequality for $\alpha_n$.

4. Due to part 2, we can assume (possibly after passing to a subsequence) that $m_n\to\tilde m$ for some $\tilde m$. If $\sigma_n$ is bounded, then (possibly after passing to a subsequence)  $\sigma_n\to\tilde\sigma$ and $\tilde\sigma>0$ due to part 3. Then by Theorem~3.1 in~\cite{GurHannesGCP}, $\tilde m=m_g=0$. Furthermore, since $\sigma_n\to\tilde\sigma>0$, it follows from~\eqref{eqDKsigmaPureZero} that $\alpha_n\to\tilde\alpha>0$. Thus, $\tilde\sigma,\tilde\alpha>0$ solve the equations~\eqref{eqDKmPureZero}, \eqref{eqDKsigmaPureZero} with $m=0$. However, by Theorem~3.2, item~(c) in~\cite{GurHannesGCP}, the system of these two equations has no solution for $\sigma,\alpha>0$. Therefore, $\sigma_n,\alpha_n\to\infty$, and for sufficiently large $n$, they enter a region where, by part~1, the solution  $(\ep_n,m_n,\alpha_n,\sigma_n)$ is unique.

\section{Proof of Theorem~$4.1$}\label{secMeanExpCloseProof}

For the proof of Theorem~4.1, we need two auxiliary results, which are given in the next two subsections.

\subsection{Prognostic mean for any fixed $\ep$}

In this subsection, we assume that $\ep$ is fixed and is not necessarily small, and analyze how the equilibrium $m=\mp$ of Eq.~(12) gets perturbed compared with the ground truth mean $\mg$, provided that $\mo$ or $\Vo$ is large. We will see that the larger the values of $|\mo-\mg|$ or $\Vo$ are, the better the samples from $\po(y)$ are recognized as outliers and the stronger $\mp$ gets shifted towards $m_g$.

\begin{lemma}\label{thLargeMoVoSmallMep}
Let $\po(y):=\frac{1}{\sqrt{\Vo}}\tilde\po\left(\frac{y-\mo}{\sqrt{\Vo}}\right)$, where $\tilde\po(y)$ is an arbitrary distribution with zero mean and unit variance. We fix $\ep_*\in[0,1)$ and $\alpha,\sigma>0$. Then the following hold for all $\ep\in[0,\ep_*]$.
\begin{enumerate}
\item\label{thLargeMoVoSmallMep1}
If $|\mo-\mg|$ is large enough, then Eq.~(12) has an equilibrium $\mp$ in a neighborhood of $\mg$ satisfying
\begin{equation}\label{eqLargeMoVoSmallMep1}
\mp = \mg + \frac{\ep}{c_1(1-\ep)}\frac{1}{\mo-\mg} + O\left(\frac{1}{(\mo-\mg)^2}\right)
\end{equation}
as $|\mo-\mg|\to\infty,$ where
\begin{equation}\label{eqLargeMoVoSmallMep1c1}
c_1=c_1(\sigma,\Vg):=\frac{1}{\sqrt{2\pi}}\int \frac{z^2}{2\sigma+\Vg z^2}e^{-z^2/2}dz.
\end{equation}

\item\label{thLargeMoVoSmallMep2}
If $\Vo$ is large enough, then Eq.~(12) has an equilibrium $\mp$ in a neighborhood of $\mg$ satisfying, for any $\varkappa>0$,
\begin{equation}\label{eqLargeMoVoSmallMep2}
\mp = \mg + O\left(\frac{1}{\Vo^{\frac{1}{2}-\varkappa}}\right)\quad\text{as } \Vo\to\infty.
\end{equation}
\end{enumerate}
In both cases,  $O(\cdot)$ is uniform with respect to $\ep\in[0,\ep_*]$, $\mg\in\bbR$,  and $\Vg$ from bounded intervals.
\end{lemma}

\proof
Without loss of generality, assume that $\mg=0$.

\proof[Proof of item~1]
 We set $\lambda=1/\mo$ and apply the implicit function theorem to
\begin{equation}\label{eqfmlambda}
\begin{aligned}
& f(m,\lambda)  :=(1-\ep)\int \frac{y-m}{2\sigma+(y-m)^2}\pg(y)  + \ep \int \frac{\sqrt{\Vo}y+\frac{1}{\lambda}-m}{2\sigma+\left(\sqrt{\Vo}y+\frac{1}{\lambda}-m\right)^2}\tilde \po(y)dy=0.
\end{aligned}
\end{equation}
We have $f(0,0)=0$. Integrating by parts yields
$$
\begin{aligned}
\partial_m f(0,0) & = - \frac{1-\ep}{\sqrt{2\pi\Vg}} \int \partial_y\left(\frac{y}{2\sigma+y^2}\right)e^{-y^2/(2\Vg)}dy
\\
& = -\frac{1-\ep}{\Vg\sqrt{2\pi\Vg}} \int \frac{y^2}{2\sigma+y^2} e^{-y^2/(2\Vg)}dy \\
& =-(1-\ep)c_1,
\end{aligned}
$$
where $c_1$ is defined in~\eqref{eqLargeMoVoSmallMep1c1}. Further, $\partial_\lambda f(0,0) = \varepsilon$. Hence, there is a neighborhood of $(0,0)$ in which Eq.~\eqref{eqfmlambda} has a unique root $\mp=\mp(\lambda)$ for each fixed $\lambda$, and
$$
\frac{d \mp}{d\lambda}\Big|_{\lambda=0} = \frac{\ep}{c_1(1-\ep)}.
$$
Finally, one can check that the second derivatives of $f(m,\lambda)$ are continuous in a neighborhood of $(0,0)$, which implies the Taylor expansion of $\mp(\lambda)$ equivalent to~\eqref{eqLargeMoVoSmallMep1}.
\endproof

\proof[Proof of item~2]
We fix an arbitrary $\varkappa>0$ and set $\delta=\Vo^{\frac{-1}{4+2\varkappa}}$, so that $\delta^{2+\varkappa}=\Vo^{-1/2}$. We will apply the implicit function theorem to
\begin{equation}\label{eqgmlambda}
\begin{aligned}
g(m,\delta) & :=(1-\ep)\int \frac{y-m}{2\sigma+(y-m)^2}\pg(y) + \ep \int \frac{\sqrt{\Vo}y+\mo-m}{2\sigma+\left(\sqrt{\Vo}y+\mo-m\right)^2}\tilde \po(y)dy\\
& =(1-\ep)\int \frac{y-m}{2\sigma+(y-m)^2}\pg(y)  + \ep \int \frac{\frac{y}{\delta^{2+\varkappa}}+\mo-m}{2\sigma+\left(\frac{y}{\delta^{2+\varkappa}}+\mo-m\right)^2}\tilde \po(y)dy=0.
\end{aligned}
\end{equation}
We have $g(0,0)=0$, $\partial_m g(0,0)=-(1-\ep)c_1$, $\partial_\delta g(0,0)=0$. Hence, there is a neighborhood of $(0,0)$ in which Eq.~\eqref{eqgmlambda} has a unique root $\mp=\mp(\delta)$ for each fixed $\delta$, and
 $$
\frac{d \mp}{d\delta}\Big|_{\delta=0} = 0.
$$
Furthermore, one can check that the second partial derivatives of $g(m,\delta)$ are continuous in a neighborhood of $(0,0)$. Therefore,
$
\mp = O(\delta^2)
$
as $\delta\to 0$.
Since $\varkappa>0$ is arbitrary, the latter asymptotics is equivalent to~\eqref{eqLargeMoVoSmallMep2}.
\endproof

\subsection{An auxiliary algebraic relation}

For the reader's convenience, we formulate the following lemma, which is proved in~\cite[Lemma~3.1]{GurHannesGCP}

\begin{lemma}\label{lA}
  For each $\alpha>0$, the equation
\begin{equation}\label{eqA}
 \frac{2\alpha+1}{\sqrt{2\pi}}\int\frac{y^2}{2(\alpha-A) + y^2}e^{-y^2/2} \,dy - 1 =0
\end{equation}
with respect to $A$ has a unique root $A(\alpha)$. The function $A(\alpha)$ is monotone increasing from $0$ to $1$ and satisfies $\alpha-A(\alpha)>0$ for all $\alpha>0$, see  Fig.~\ref{figA}.
\end{lemma}

It implies the following corollary.

\begin{corollary}\label{corA}
  For each $\alpha,\sigma>0$, the equation
\begin{equation}\label{eqA}
 \frac{2\alpha+1}{\sqrt{2\pi}}\int\frac{y^2}{2\sigma/V + y^2}e^{-y^2/2} \,dy - 1 =0
\end{equation}
with respect to $V$ has a unique root $V=\Vp=\frac{\sigma}{\alpha-A(\alpha)}$, where $A(\alpha)$ is defined in Lemma~\ref{lA}.
\end{corollary}

\subsection{Proof of Theorem~$4.1$}

1. We set $\tilde\mp(\ep):=\mp-\mg(\ep)$ and $\tilde\mo(\ep):=\mo(\ep)-\mg(\ep)$. 

Then that $\tilde\mp$ satisfies
\begin{equation}\label{eqEqForMExpClose}
\begin{aligned}
& \frac{1}{\sqrt{2\pi}}\int \frac{\sqrt{\Vg}y-\tilde\mp}{2\sigma+(\sqrt{\Vg}y-\tilde\mp)^2}e^{-y^2/2} \\
& \quad = \ep f(\tilde\mp,\tilde\mo, \Vg,
 \Vo),
\end{aligned}
\end{equation}
where $f(\cdot)$ is uniformly bounded with respect to all its arguments. Further, $\Vg=\Vg(\ep)$ is bounded by assumption, and Eq.~\eqref{eqEqForMExpClose} with the zero right hand side has a unique solution $\tilde\mp=0$. Therefore, there exists $\tilde\mp\to 0$ as $\ep\to 0$ uniformly with respect to $\tilde\mo\in\bbR$, $\Vo>0$, and $\Vg$.

2. Due to~(13),~(18), the equilibrium $(\alpha,\sigma)$ satisfies
\begin{equation}\label{eqDKalphaPureZero1MeanExpClose}
\begin{aligned}
& G(\alpha,\sigma,\ep,\Vg,\tilde\mo,\Vo,\tilde\mp) \\
& :=  \frac{1-\ep}{\sqrt{2\pi}}\int\ln\left(1+ \frac{(\sqrt{\Vg}y  - \tilde\mp)^2}{2\sigma}\right)e^{-y^2/2}\,dy 
 + \Delta\Psi(\alpha) \\
 &  + \ep \int\ln\left(1+ \frac{(\sqrt{\Vo}y +\tilde\mo - \tilde\mp)^2}{2\sigma}\right)\tilde\po(y)\,dy =0,
\end{aligned}
\end{equation}
\begin{equation}\label{eqDKsigmaPureZero1MeanExpClose}
\begin{aligned}
& H(\alpha,\sigma,\ep,\Vg,\tilde\mo,\Vo,\tilde\mp) \\
 & := \frac{(1-\ep)(2\alpha+1)}{\sqrt{2\pi}}\int\frac{(\sqrt{\Vg}y-\tilde\mp)^2}{2\sigma + (\sqrt{\Vg}y - \tilde\mp)^2} e^{-y^2/2} \,dy\\
&  + \ep(2\alpha+1) \int\frac{(\sqrt{\Vo}y+\tilde\mo - \tilde\mp)^2}{2\sigma + (\sqrt{\Vo}y+\tilde\mo - \tilde\mp)^2} \tilde\po(y) \,dy
-1  =0.
\end{aligned}
\end{equation}
Note that the functions $G$ and $H$ coincide with those in~\eqref{eqDKalphaPureZero} and~\eqref{eqDKsigmaPureZero} (the latter up to a sign), but here we explicitly indicate their dependence on $\Vg$, $\tilde\mo$, $\Vo$, and $\tilde\mp$.

Using Corollary~\ref{corA} and the fact that $\tilde\mp\to 0$, we can pass to the limit in~\eqref{eqDKsigmaPureZero1MeanExpClose} as $\ep\to 0$, and we see that $\Vg(\ep)\to\Vp$. Hence, passing to the limit in~\eqref{eqDKalphaPureZero1MeanExpClose}, we have
$$
\ep \int\ln\left(1+ \frac{(\sqrt{\Vo}y +\tilde\mo - \tilde\mp)^2}{2\sigma}\right)\tilde\po(y)\,dy = b_0 + o(1),
$$
where
\begin{equation}\label{eqb0alpha}
\begin{aligned}
 b_0 & = b_0(\alpha):=  - \frac{1}{\sqrt{2\pi}}\int\ln\left(1+ \frac{\Vp y^2}{2\sigma}\right)e^{-y^2/2}\,dy
   - \Delta\Psi(\alpha),
\end{aligned}
\end{equation}
Since  $\Vo(\ep)$ or $\tilde\mo(\ep)$ are bounded by assumption, we obtain $\tilde\mo=e^{\frac{b_0+o(1)}{2\ep}}$ or $\Vo=e^{\frac{b_0+o(1)}{\ep}}$, respectively. Combining this with Lemma~\ref{thLargeMoVoSmallMep} concludes the proof.

\section{Proof of Theorem~$4.2$}\label{secVpDiffProof}

%

In the formulation of Theorem~4.2, we use the constant
\begin{equation}\label{eqb2alpha}
  b  = b(\alpha) := 2\alpha\left(  \frac{2\alpha+1}{\sqrt{2\pi}}\int\frac{y^2 2\sigma/\Vp}{(2\sigma/\Vp + y^2)^2}e^{-y^2/2} \,dy  \right)^{-1}.
\end{equation}
In the proof, we will also need the constant
\begin{equation}\label{eqb1alpha}
\begin{aligned}
  b_1  =b_1(\alpha)  :=  \frac{1}{\sqrt{2\pi}}\int\ln\left(1+ \frac{\Vp y^2}{2\sigma}\right)e^{-y^2/2}\,dy
 + \frac{1}{\sqrt{2\pi}}\int \left(\frac{\frac{\Vp b_2 y^2}{2\sigma}}{1+\frac{\Vp y^2}{2\sigma}}\right)e^{-y^2/2}\,dy.
\end{aligned}
\end{equation}
Note that after substituting $\Vp$ given by~(6), the variable $\sigma$ cancels. Thus $b$ and $b_1$ are indeed functions of $\alpha$ only, with $\lim\limits_{\alpha\to 0}b(\alpha)=2$. 

We will prove the theorem under the assumption that $\Vo$ does not depend on $\ep$. The case where $\mo$ does not depend on $\ep$ is analogous.

Without loss of generality, assume that $m=\mg=0$.
Due to~(13),~(18), the equilibrium $(\alpha,\sigma)$ satisfies
\begin{equation}\label{eqDKalphaPureZero1}
\begin{aligned}
& G(\alpha,\sigma,\ep,\Vg,\mo,\Vo) \\
 & :=  \frac{1-\ep}{\sqrt{2\pi}}\int\ln\left(1+ \frac{\Vg y^2}{2\sigma}\right)e^{-y^2/2}\,dy
  + \Delta\Psi(\alpha) \\
 &  + \ep \int\ln\left(1+ \frac{(\sqrt{\Vo}y+\mo)^2}{2\sigma}\right)\tilde\po(y)\,dy =0,
\end{aligned}
\end{equation}
\begin{equation}\label{eqDKsigmaPureZero1}
\begin{aligned}
& H(\alpha,\sigma,\ep,\Vg,\mo,\Vo) \\
&   := \frac{(1-\ep)(2\alpha+1)}{\sqrt{2\pi}}\int\frac{y^2}{2\sigma/\Vg + y^2}e^{-y^2/2} \,dy\\
&  + \ep(2\alpha+1) \int\frac{(\sqrt{\Vo}y+\mo)^2}{2\sigma + (\sqrt{\Vo}y+\mo)^2} \tilde\po(y) \,dy
-1  =0.
\end{aligned}
\end{equation}
Note that the functions $G$ and $H$ are the same as in~\eqref{eqDKalphaPureZero1MeanExpClose} and~\eqref{eqDKsigmaPureZero1MeanExpClose}, but we omit the dependence on $\mp$, which is assumed to coincide with $\mg$.


We will show that one can find unique roots $\Vg=\Vg(\ep)$ and $\mo=\mo(\ep)$ of Eq.~\eqref{eqDKalphaPureZero1} and~\eqref{eqDKsigmaPureZero1} as  functions of $\ep$ (and the other parameters) and determine their asymptotics, provided $\ep$ is small. First, assume that $\Vg=\Vg(\ep)$ and $\mo=\mo(\ep)$ exist for all sufficiently small $\ep$. Then $\Vg(\ep)$ is bounded as $\ep\to 0$. Otherwise, passing to the limit in~\eqref{eqDKsigmaPureZero1}, we would obtain $2\alpha=0$. Furthermore, it is bounded away from zero. Otherwise,  passing to the limit in~\eqref{eqDKsigmaPureZero1}, we would obtain $-1=0$. Thus, in what follows, it suffices to consider $\Vg$ from a bounded interval separated from zero.

We introduce the variable $\mu$ instead of $\mo$ such that $\mo=\mo(\ep,\mu)=(2\sigma)^{1/2}e^{b_1/2}e^{b_0/(2\ep)}(1+\mu)$ and prove existence of $\mu(\ep),\Vg(\ep)$. Here $b_0$ is given by~\eqref{eqb0alpha} and $b_1$ by~\eqref{eqb1alpha}.

 First, we solve Eq.~\eqref{eqDKsigmaPureZero1} for $\Vg=\Vg(\ep,\mu)$. Consider the function $\tilde H(\ep,\mu,\Vg):=H(\alpha,\sigma,\ep,\Vg,\mo(\ep,\mu))$.
Note that there is $\ep_1\in(0,1)$ independent of $\mu$ such that for all $\ep\in[0,\ep_1]$ and $\mu\in\bbR$,
$$
\lim\limits_{V_g\to 0} \tilde H(\ep,\mu,\Vg) <0, \qquad \lim\limits_{V_g\to \infty} \tilde H(\ep,\mu,\Vg) >0,
$$
and $\tilde H(\ep,\mu,\Vg)$ is monotone with respect to $\Vg$. Hence, Eq.~\eqref{eqDKsigmaPureZero1}  has a unique root $\Vg=\Vg(\ep,\mu)$ for all $\ep\in[0,\ep_1]$ and $\mu$, and, due to Corollary~\ref{corA}, $\Vg(\ep,\mu)=\Vp+o(1)$, where $o(1)$ is uniform with respect to all $\mu\in\bbR$. The partial derivatives of $\tilde H$ with respect to all its arguments are continuous for all $\ep\in[0,\ep_1]$, and $\mu,\Vg$. Furthermore, as $\ep\to 0$, we have
$$
\begin{aligned}
& \partial_{\Vg}\tilde H  = \frac{(1-\ep)(2\alpha+1)}{\sqrt{2\pi}}\int\frac{y^2 2\sigma/\Vg^2 }{(2\sigma/\Vg + y^2)^2}e^{-y^2/2} \,dy,\\
& \partial_{\ep}\tilde H  = -\frac{2\alpha+1}{\sqrt{2\pi}}\int\frac{y^2}{2\sigma/\Vg + y^2}e^{-y^2/2} \,dy \\
& + (2\alpha+1)\left(1-\int\frac{2\sigma}{2\sigma + (\sqrt{\Vo}y+\mo)^2} \tilde\po(y) \,dy\right) \\
& - \frac{(2\alpha+1)\mo}{\ep}\int\frac{\frac{\sqrt{\Vo}y+\mo}{\sigma}}{\left(1+ \frac{(\sqrt{\Vo}y+\mo)^2}{2\sigma}\right)^2}\tilde\po(y)\,dy,\\
& \partial_\mu\tilde H   = \frac{\ep}{1+\mu} \int\frac{\frac{\sqrt{\Vo}y+\mo}{\sigma}}{\left(1+ \frac{(\sqrt{\Vo}y+\mo)^2}{2\sigma}\right)^2}\tilde\po(y)\,dy.
\end{aligned}
$$
Hence, by the implicit function theorem,  $\Vg(\ep,\mu)$ is continuously differentiable with respect to $\ep,\mu$ for all $\ep\in[0,\ep_1]$ and $\mu\in\bbR$. In particular,
\begin{equation}\label{eqPartialVgep}
\partial_\ep\Vg|_{\ep=0} =  -\Vp b,
\end{equation}
where $b$ is defined in Eq.~\eqref{eqb2alpha}.

We substitute $\mo=\mo(\ep,\mu)$ and $\Vg=\Vg(\ep,\mu)$ into~\eqref{eqDKalphaPureZero1}, and obtain the equation
\begin{equation}\label{eqGtilde}
(1-\ep)\Gg(\ep,\mu) + \ep \Go(\ep,\mu) + \Delta\Psi(\alpha) = 0,
\end{equation}
where
\begin{equation}\label{eqGgGo2}
\begin{aligned}
\Gg(\ep,\mu) &:= \frac{1}{\sqrt{2\pi}}\int\ln\left(1+ \frac{\Vg(\ep,\mu) y^2}{2\sigma}\right)e^{-y^2/2}\,dy,\\
\Go(\ep,\mu) &:= \int\ln\left(1+ \frac{(\sqrt{\Vo}y+\mo(\ep,\mu))^2}{2\sigma}\right)\tilde\po(y)\,dy.
\end{aligned}
\end{equation}
Note that
$$
\begin{aligned}
& \Vg(\ep,\mu)  =\Vg(\ep,0)+ \ep^2 f_1(\ep,\mu) \\
& =  \Vp + \ep \partial_\ep\Vg|_{(\ep,\mu)=(0,0)} + \ep^2 (f_1(\ep,\mu)+f_2(\ep))\\
& = \Vp - \ep \Vp b + \ep^2 (f_1(\ep,\mu)+f_2(\ep)),
\end{aligned}
$$
where
$$
\begin{aligned}
& f_1(\ep,\mu)=\ep^{-2}\int_{0}^{\mu}\partial_{\mu'}\Vg(\ep,\mu')\,d\mu',\\
& f_2(\ep)=\ep^{-2}\int_{0}^{\ep}\int_{0}^{\ep'}\partial^2_{\ep''}\Vg(\ep'',0)\,d\ep'' d\ep',
\end{aligned}
$$
and $b$ is defined in~(27).
Therefore,
\begin{equation}\label{eqGgGo3}
\begin{aligned}
& (1-\ep)\Gg(\ep,\mu)   = \frac{1-\ep}{\sqrt{2\pi}} \int\ln\left(1+ \frac{\Vp y^2}{2\sigma}\right)e^{-y^2/2}\,dy \\
& + \frac{1-\ep}{\sqrt{2\pi}}\int\ln\biggl(1+ 
  \frac{(-\ep\Vp b + \ep^2 (f_1(\ep,\mu)+f_2(\ep)))y^2/\sigma}{1+ \frac{\Vp y^2}{2\sigma}}\biggr) e^{-y^2/2}\,dy \\
& = -b_0-\Delta\Psi(\alpha) - \ep b_1 + \ep^2 f_3(\ep,\mu).
\end{aligned}
\end{equation}
where $b_0$ and $b_1$ are defined in~\eqref{eqb0alpha} and~\eqref{eqb1alpha} and $f_3,\partial_\mu f_3,\partial_\ep f_3$ are bounded and continuous for $\ep\in[0,\ep_1]$ and all $\mu\in\bbR$.

Further,
\begin{equation}\label{eqGgGo4}
\ep\Go(\ep,\mu) = \ep f_4(\ep,\mu)  + \ep b_1 + b_0,
\end{equation}
where
$$
\begin{aligned}
& f_4(\ep,\mu):= \int\ln\biggl( e^{-b_1-\frac{b_0}{\ep}}  + \left(y(2\sigma)^{-1/2}e^{-\frac{b_1}{2}-\frac{b_0}{2\ep}} +1 + \mu\right)^2 \biggr)e^{-y^2/2}dy.
\end{aligned}
$$

Combining~\eqref{eqGtilde}--\eqref{eqGgGo4}, we see that, for $\ep>0$, Eq.~\eqref{eqGtilde} is equivalent to
\begin{equation}\label{eqGhat}
\hat G(\ep,\mu):=f_4(\ep,\mu) + \ep f_3(\ep,\mu)=0.
\end{equation}
We have $\hat G(0,0)=0$, the partial derivatives of $\hat G$ are continuous for all $\ep\in[0,\ep_1]$ and $\mu\in\bbR$, and $\partial_\mu \hat G(0,0)>0$. Hence, by the implicit function theorem, there exist small $\ep_*>0$ and $\mu_*\in\bbR$ such that Eq.~\eqref{eqGhat} has a unique solution $\mu=\mu(\ep)$ for all $\ep\in[0,\ep_*]$, $\mu\in[-\mu_*,\mu_*]$. This solution is continuously differentiable in a neighborhood of the origin. Similarly, there is a unique solution $\mu=\mu(\ep)$ for all $\ep\in[0,\ep_*]$, $\mu\in[-2-\mu_*,-2+\mu_*]$. To prove that there are no solutions outside of these two $\mu$-regions, one can show that $f_4(\ep,\mu)$ is monotone decreasing for $\mu\in(-\infty,-1+\mu_1(\ep))$, monotone increasing for $\mu\in(-1+\mu_1(\ep),\infty)$, and $\mu_1(\ep)\to 0$.
This proves~(29). Applying the chain rule to $\Vg(\ep,\mu(\ep))$ also  yields~(28).

%
%
%
%

\section{The RMSE($n$) curves}

\begin{figure*}[t]
\vskip 0.2in
\begin{center}
	\begin{minipage}{0.3\textwidth}
	   \includegraphics[width=\textwidth]{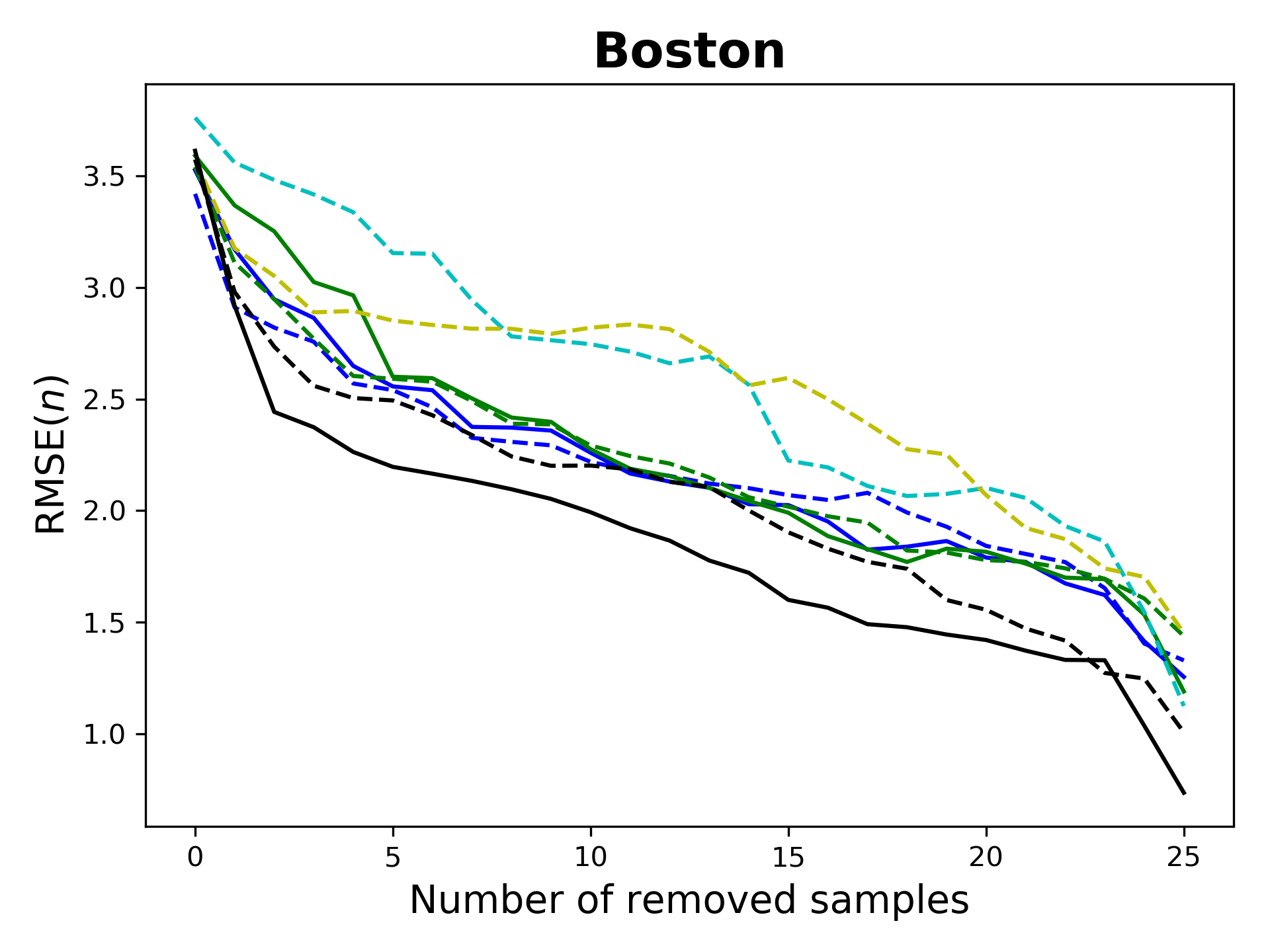}
	\end{minipage}
\hfill
	\begin{minipage}{0.3\textwidth}
       \includegraphics[width=\textwidth]{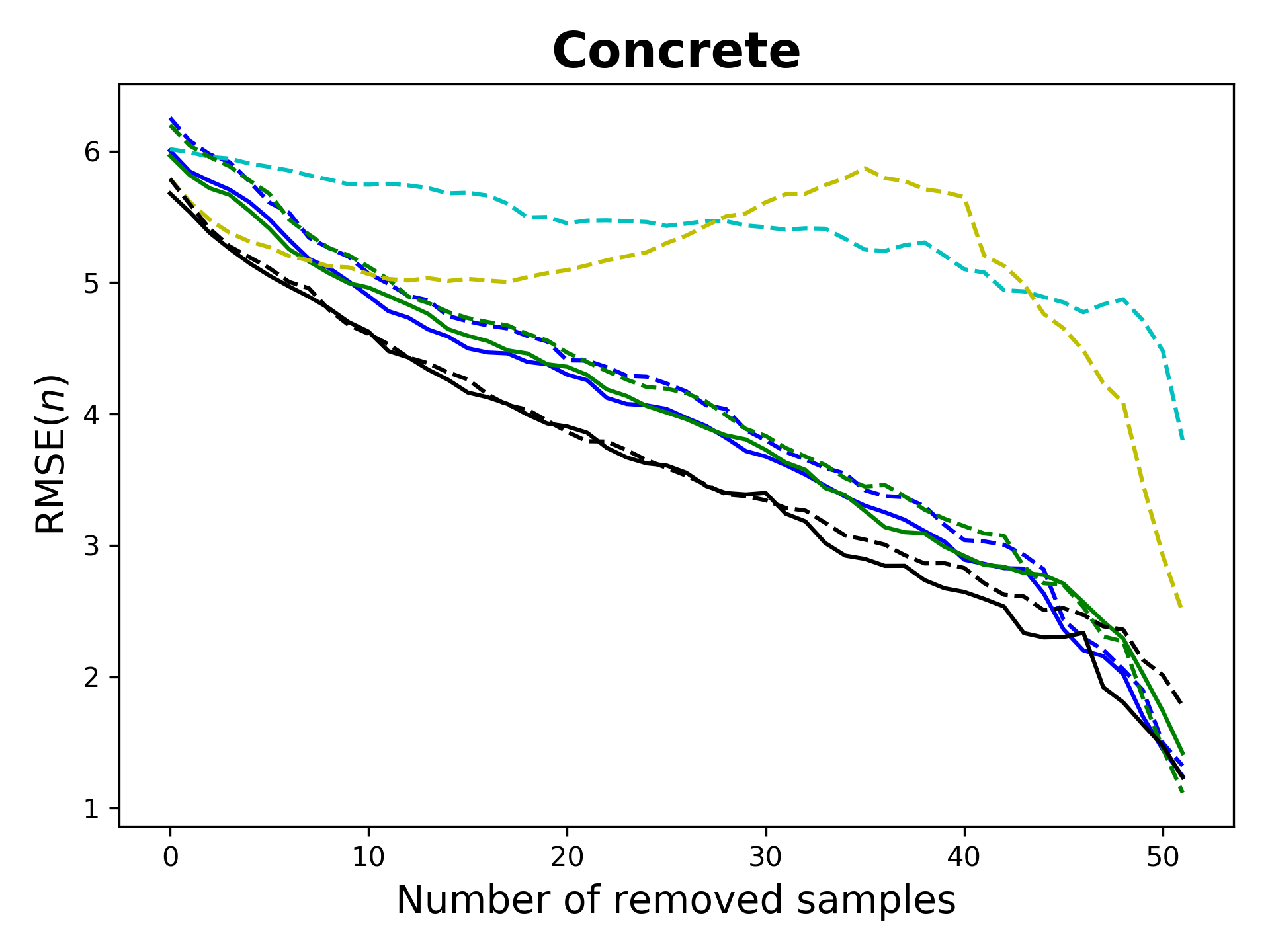}
	\end{minipage}
\hfill
	\begin{minipage}{0.3\textwidth}
	   \includegraphics[width=0.45\textwidth]{legends.png}
	\end{minipage}

	\begin{minipage}{0.3\textwidth}
       \includegraphics[width=\textwidth]{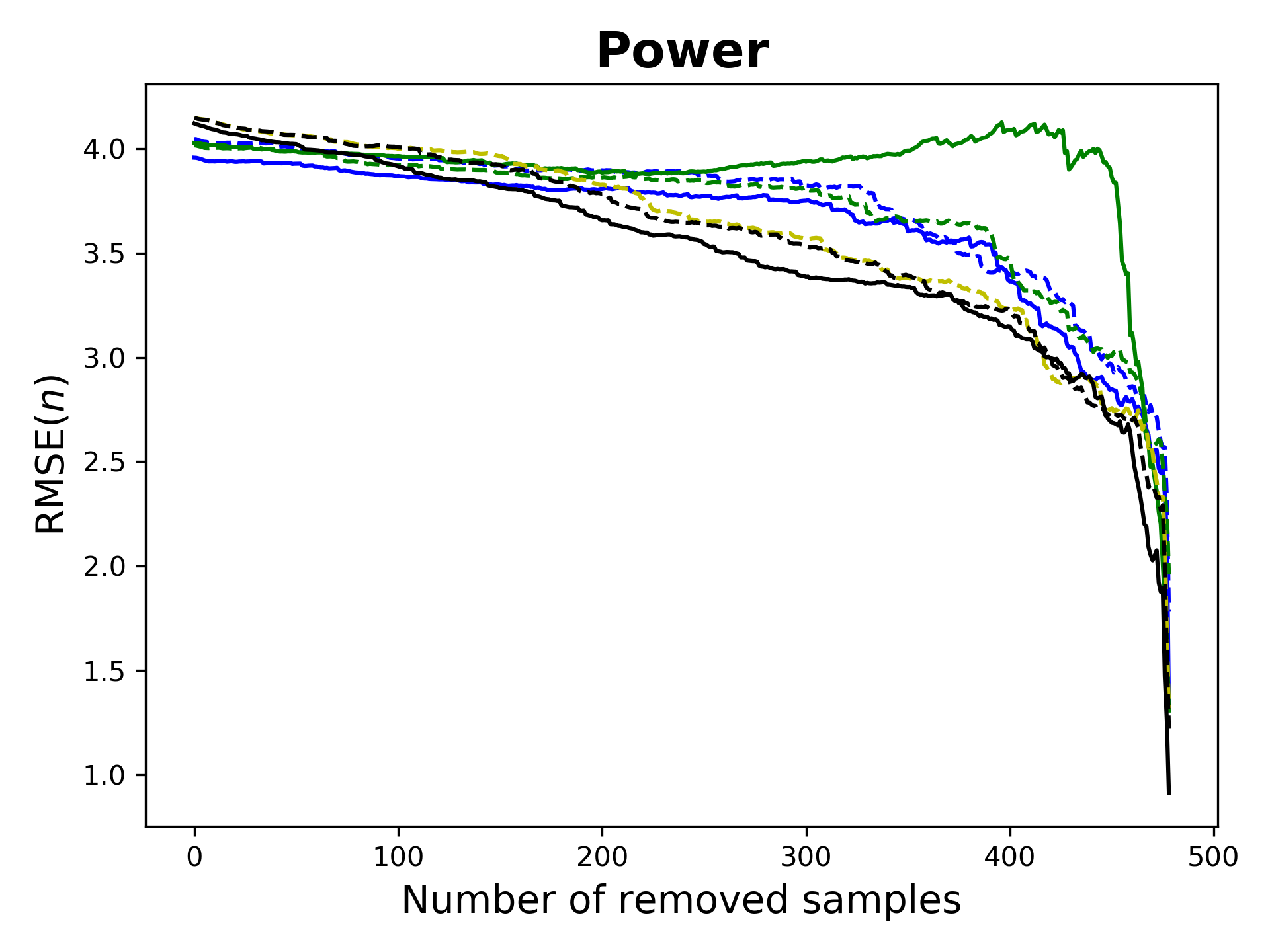}
	\end{minipage}
\hfill
	\begin{minipage}{0.3\textwidth}
       \includegraphics[width=\textwidth]{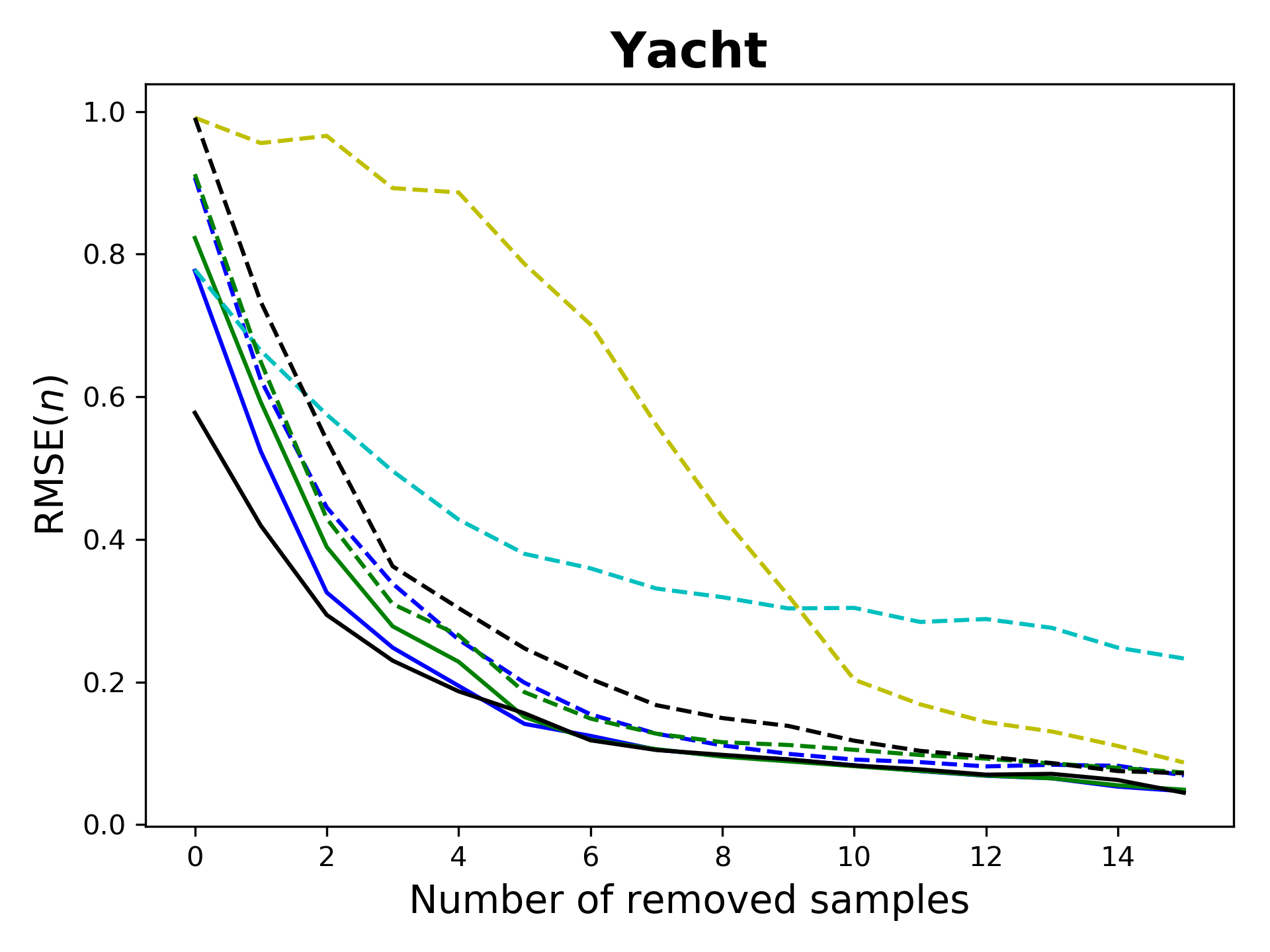}
	\end{minipage}
\hfill
	\begin{minipage}{0.3\textwidth}
       \includegraphics[width=\textwidth]{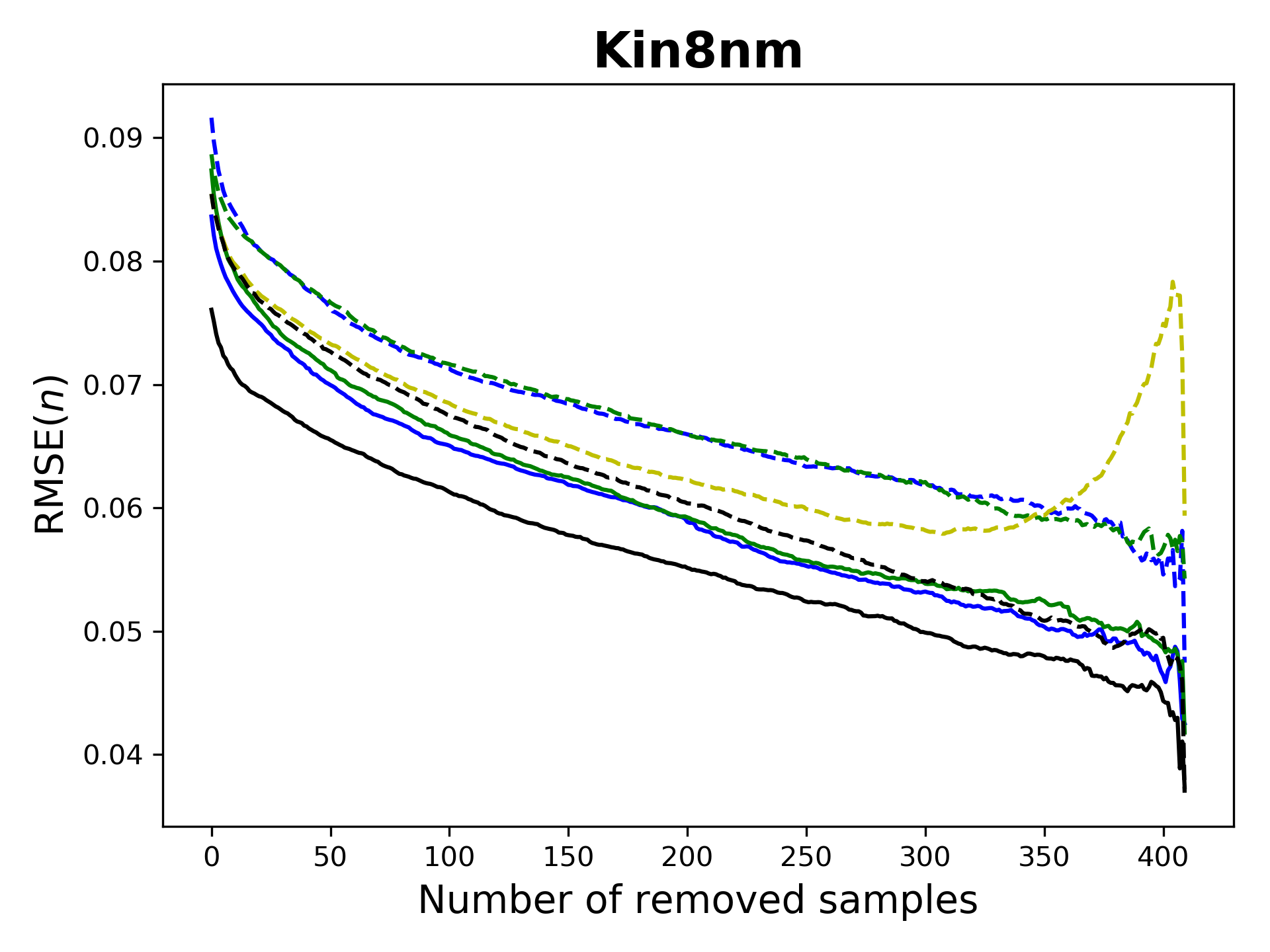}
	\end{minipage}
\caption{RMSE($n$) curves for different methods and data sets from Sec.~6.3. The training sets are contaminated by 5\% of outlier.}\label{figRemovedSamples}
\end{center}
\vskip -0.2in
\end{figure*}

Figure~\ref{figRemovedSamples} shows the RMSE($n$) curves for different methods and data sets from Sec.~6.3, fitted on training sets contaminated by 5\% of outlier. We see that (possibly after removing a small number of samples for which EnsGCP predicts a high variance) its RMSE is significantly better than the respective RMSE of the other methods, see the curves ${\rm RMSE}(n)$ in Fig.~5 in the supplement.

\section{Architectures and hyperparameters}\label{appendixHyperparameters}

\subsection{Architectures}
We use one-hidden layer networks with 50 ReLU
nonlinearities for the Beta, Gamma, and GCP. Whenever a method uses several quantities (e.g., the mean and variance in the Beta and Gamma, or $m,\nu,\alpha,\beta$ in the GCP), we approximate each quantity by a separate network. For regularization in non-Bayesian methods, we use a dropout layer between the hidden layer and the output unit. Our approach is directly applicable to neural networks of any depth and structure, however we keep one hidden layer for the compatibility of our validation with~\cite{HernandezLobato15,Lakshminarayanan17,GurHannesLU}. For {Beta$_{\rm Bayes}$}, we used the architecture from the authors' code\footnote{\url{https://github.com/futoshi-futami/Robust_VI}}.

\subsection{Hyperparameters}

When we fit all the methods except the Beta$_{\rm Bayes}$, we first contaminate the training set by outliers and then normalize it such that the input features and the targets
have zero mean and unit variance. For the Beta$_{\rm Bayes}$, significantly better results were achieved without normalizing the targets\footnote{Note that the authors in~\cite{Futami2017} used a different protocol, namely they normalized the noncontaminated training set and then added outliers to it.}.

For the Beta, Gamma, and GCP, we used minibatch 5 on Boston, Concrete, and Yacht, and minibatch 10 on Power and Kin8nm. We used Adam (with $\beta_1=0.9$, $\beta_2=0.999$) optimizer for fitting, and performed a grid search for the learning rate in the range $\{0.00002, 0.00005, 0.0001, 0.0002, 0.0007, 0.001, 0.005\}$ and for the dropout rate in the range $\{0,0.1,0.2,0.3,0.4\}$. For the Beta and Gamma, we optimized for the learning rate with the fixed parameters $\beta=1$ and $\gamma=1$, respectively. After that,
we additionally performed a grid search for $\beta$ and $\gamma$ in the range $\{0.1,0.2,\dots,1\}$. We observed that changing the learning rate for the newly found values of $\beta$ and $\gamma$ did not significantly improve the results.
All the grid searches was performed for training data sets with 5\% of outliers and evaluated on the noncontaminated test sets. The optimized parameters are given in Table~\ref{tableHyperparametersBetaGammaGCP}. For the ensemble methods, we used the hyperparameters that were optimal for the respective non-ensemble methods, but with the half dropout rate.  For {Beta$_{\rm Bayes}$}, we used the architecture, the default settings and the optimizer based on the Edward library~\cite{tran2016edward} as in the authors' code\footnote{\url{https://github.com/futoshi-futami/Robust_VI}}, and we performe a grid search for the parameter $\beta=0.1,0.2,\dots,1$ and the standard deviation of the likelihood $\sigma=0.1, 0.5, 1, 2, 4, 6, 8, 10.$

\begin{table*}[t]
\caption{ Learning rate {\bf LR}, dropout rate {\bf D}, and the number of epochs {\bf NE} for the Beta, Gamma and GCP-based methods.}\label{tableHyperparametersBetaGammaGCP}
\vskip 0.15in
\resizebox{\textwidth}{!}{%
\begin{tabular}{lcccc}
{} & {\bf Boston} &  {} & {}\\
\toprule
{} &  {\bf LR} & {\bf D} & {\bf NE}  \\
\midrule
{\bf Beta, Gamma}                       &            0.00002  & 0.4 & 2500   \\
{\bf GCP}                       &            0.0001  & 0.3 & 700   \\
\bottomrule
\end{tabular}
\
\begin{tabular}{cccc}
 {\bf Concrete} & {} & {} & {}\\
\toprule
  {\bf LR} & {\bf D} & {\bf NE}  \\
\midrule
      0.00001  & 0.1 & 2500  \\
      0.0001  & 0.1 & 1000  \\
\bottomrule
\end{tabular}
\
\begin{tabular}{cccc}
 {\bf Power} & {} & {} & {}\\
\toprule
  {\bf LR} & {\bf D} & {\bf NE}  \\
\midrule
      0.0001 & 0 & 400 \\
      0.00005 & 0 & 150 \\
\bottomrule
\end{tabular}
\
\begin{tabular}{lcccc}
{} & {\bf Yacht} &  {} & {}\\
\toprule
  {\bf LR} & {\bf D} & {\bf NE}  \\
\midrule
            0.0001 & 0.1 & 2500   \\
            0.001 & 0.1 & 1000   \\
\bottomrule
\end{tabular}

\begin{tabular}{cccc}
 {\bf Kin8nm} &  {} & {}\\
\toprule
 {\bf LR} & {\bf D} & {\bf NE}  \\
\midrule
     0.0001 & 0 & 400  \\
     0.0007 & 0 & 250  \\
\bottomrule
\end{tabular}
}
\vskip -0.1in
\end{table*}

\begin{table}[t]
\caption{Optimal values of $\beta$ and $\gamma$ for the Beta and Gamma methods.}\label{tableHyperparametersBetaGamma}
\begin{center}
\begin{small}
\begin{sc}
\resizebox{0.5\textwidth}{!}{%
\begin{tabular}{lc}
{} & {\bf Boston} \\
\toprule
{$\beta$}                       &            0.2     \\
{$\gamma$}                      &            0.4     \\
\bottomrule
\end{tabular}
\
\begin{tabular}{c}
 {\bf Concrete}\\
\toprule
      0.6  \\
      0.6 \\
\bottomrule
\end{tabular}
\
\begin{tabular}{c}
 {\bf Power}\\
\toprule
      0.1  \\
      0.1 \\
\bottomrule
\end{tabular}
\
\begin{tabular}{c}
 {\bf Yacht}\\
\toprule
      0.4  \\
      0.4 \\
\bottomrule
\end{tabular}

\begin{tabular}{c}
 {\bf Kin8nm}\\
\toprule
      0.2  \\
      0.2 \\
\bottomrule
\end{tabular}
}
\end{sc}
\end{small}
\end{center}
\end{table}

\begin{table}[t]
\caption{Optimal values of $\beta$ and the standard deviation $\sigma$ of the likelihood for the Beta$_{\rm Bayes}$ method. Symbol $*$ indicates that we were not able to fine tune the parameters of the Beta$_{\rm Bayes}$ to obtain reasonable predictions for Power and Kin8nm data sets. Note that the authors in~\cite{Futami2017} used a protocol for fitting Beta$_{\rm Bayes}$ different from ours. Unlike us, they first normalized the noncontaminated training set and then added outliers to~it}\label{tableHyperparametersBetaBayes}
\begin{center}
\begin{small}
\begin{sc}
\resizebox{0.5\textwidth}{!}{%
\begin{tabular}{lc}
{} & {\bf Boston} \\
\toprule
{$\beta$}                       &            0.1     \\
{$\sigma$}                      &            1     \\
\bottomrule
\end{tabular}
\
\begin{tabular}{c}
 {\bf Concrete}\\
\toprule
      0.1  \\
      4 \\
\bottomrule
\end{tabular}
\
\begin{tabular}{c}
 {\bf Power}\\
\toprule
      *  \\
      * \\
\bottomrule
\end{tabular}
\
\begin{tabular}{c}
 {\bf Yacht}\\
\toprule
      0.1  \\
      0.5 \\
\bottomrule
\end{tabular}

\begin{tabular}{c}
 {\bf Kin8nm}\\
\toprule
      *  \\
      * \\
\bottomrule
\end{tabular}
}
\end{sc}
\end{small}
\end{center}
\end{table}

%

\end{document}